\RequirePackage{fix-cm}
\documentclass{styles/svproc}
\usepackage{amsmath}
\usepackage[font={scriptsize}]{caption}
\usepackage{amssymb} % assumes amsmath package installed
\usepackage{amsfonts}
\usepackage{mathtools}
\usepackage{algorithmic}
\usepackage{algorithm}
\usepackage{array}
\usepackage{textcomp}
\usepackage{stfloats}
\usepackage{url}
\usepackage{verbatim}
\usepackage{graphicx}
\usepackage{cite}
\usepackage[inkscapeformat=png]{svg}
\usepackage{subcaption}
\usepackage[font={scriptsize}]{caption}
\usepackage[section]{placeins}
\usepackage{float}
\usepackage{booktabs}
\usepackage{makecell}
\usepackage{blindtext}
\usepackage{comment}
\usepackage{mathbbol}
\usepackage{tablefootnote}
\usepackage{footnotehyper}
\usepackage[colorlinks,linkcolor=blue]{hyperref}
\graphicspath{{./figures/}}

\usepackage{balance}
\usepackage{multirow}

\usepackage{framed}
\usepackage{wrapfig}
\usepackage{import}
\usepackage{arydshln}
\usepackage{multicol}
\usepackage{hyperref}
\hypersetup{
 colorlinks=true,
 linkcolor=black,
 citecolor=black,
 filecolor=magenta, 
 urlcolor=blue,
 }

\newcommand{\reals}{\mathbb{R}}

\newcommand{\URL}[1]{{\color{blue}\textit{Supplemental Video:} #1}}

\usepackage{lmodern} 
\usepackage{epsfig} % for postscript graphics files

\usepackage{url}

\begin{document}
\mainmatter              % start of a contribution
\title{\LARGE \bf Decentralized Adaptive Aerospace Transportation of Unknown Loads Using A Team of Robots}

\titlerunning{Decentralized aerospace transport}

\author{Longsen Gao\inst{1} \and Kevin Aubert\inst{1}  \and David Saldaña\inst{3} \and Claus Danielson\inst{2} \and \\
Rafael Fierro\inst{1}}

\authorrunning{Longsen Gao et al.}

\institute{Department of Electrical and Computer Engineering \\
  %  The University of New Mexico, Albuquerque NM 87131, USA. \\
  %  \email{lgao1,kevinaubert1,rfierro@unm.edu}
    \and
    Department of Mechanical Engineering \\
    The University of New Mexico, Albuquerque NM 87131, USA.\\
    \email{\{lgao1,kevinaubert1,rfierro,cdanielson\}}@unm.edu
    \and
    Department of Computer Science \& Engineering \\
    Lehigh University, Bethlehem PA 18015, USA.\\
    \email{saldana@lehigh.edu}
    }

\maketitle              % typeset the title of the contribution

\begin{abstract}
Transportation missions in aerospace are limited to the capability of each robot and the properties of the object being transported, such as mass, inertia, and grasping points.
We present a novel decentralized adaptive controller design for multiple robots that can be implemented in different kinds of aerospace robots. Moreover, 
our controller adapts to unknown objects in different gravity environments. 
We validate our method in an aerial scenario using multiple fully actuated hexarotors with grasping capabilities, and a space scenario using a group of space tugs. In both cases, the robots cooperatively transport a payload along desired three-dimensional trajectories. We demonstrate that our method can adapt to unexpected changes, including the loss of robots during the transportation mission.\\
\URL{\href{http://tiny.cc/dars2024unm}{http://tiny.cc/dars2024unm}} 
\end{abstract}
\section{Introduction}
The payload capacity of an aerospace robot, such as an Uncrewed Aerial Vehicle (UAV) or a space tug, presents a fundamental constraint that limits its utility across various domains, notably in the construction and transportation of substantial payloads. Augmenting the payload capacity of an aerospace robot involves intricate mechanical redesigns, which are often resource-intensive. An alternative approach that leverages the collaborative efforts of multiple aerospace robots for transporting and manipulating target payload emerges as a cost-effective and promising solution. Multiple aerospace robots for cooperative manipulation and transportation considering the dynamic coupling between the individual subsystems have been widely studied in both space~\cite{gao2024adaptive,down2023adaptive,gao2023autonomous} and aerial environments~\cite{hert2023mrs,saldana2018modquad,salinas2023unified} in recent decade.

\begin{figure}[H]
\centering
\captionsetup[sub]{font=scriptsize,labelfont={sf},oneside}
\begin{subfigure}{.6\textwidth}
  \centering
  \includegraphics[width=0.87\linewidth]{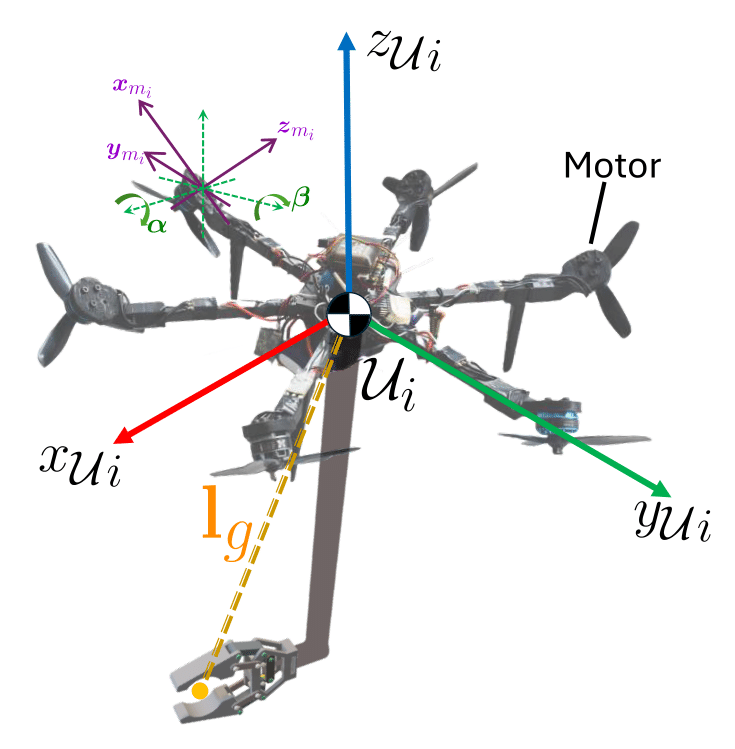}
  \caption{ }
  \label{fig_drone}
\end{subfigure}%
\begin{subfigure}{.4\textwidth}
  \centering
  \includegraphics[width=0.975\linewidth]{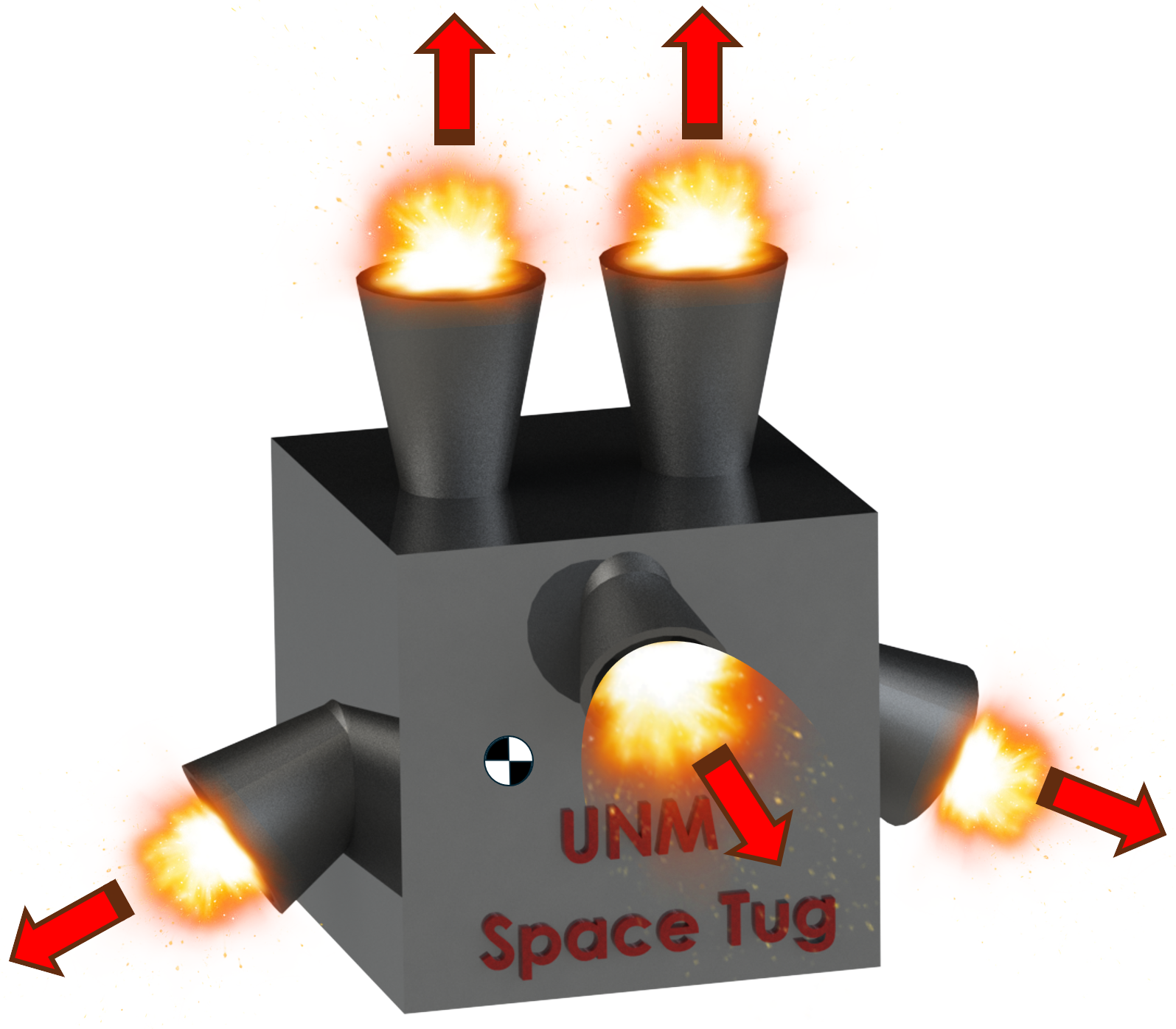}
  \vspace{1.075cm}
   \caption{ }
  \label{fig:tug}
\end{subfigure}
\caption{(a) Fully actuated hexarotor UAV with a grasping mechanism. (b) Space tug with 6 rockets around its sides. The red arrows indicate the propulsion direction of
each rocket.}
\end{figure}

\vspace{-0.5pt}
The domain of cooperative aerial transportation facilitated through tethers has been the subject of extensive research~\cite{cardona2021adaptive,sun2023nonlinear,wang2024enhancing,10611458,wang2023model}. These investigations underscore the potential of leveraging contact forces in cooperative transportation tasks, highlighting the significance of designing aerial vehicles capable of synchronized complex maneuvers. However, manipulating a tether-suspended object using multiple aerospace robots requires considering the feasibility of each robot's position with the contact force direction to avoid damage to the whole system, increasing the complexity of the internal dynamics analysis. In this case, recent research works in~\cite{bosio2023automated,mu2019universal,barawkar2024decentralized,chaikalis2023modular} learn rigid attachments for transportation purposes using a group of drones that simplify internal dynamics analysis and improve maneuverability. However, the rigid attachment scheme in those studies assumes the payload and aerospace robots are firmly connected through a non-detachable mechanic structure, decreasing the aerospace robots' dexterity. The work in~\cite{mellinger2013cooperative} studies a group of drones installed grippers in cooperative payload stabilization task. The gripper aims to penetrate the surface to hold the object, introducing the hidden risk of damage and limiting the scope of the object material. An inflight self-disassembly gripper design in ~\cite{icra18gripper} uses a self-adaptation scheme by self-reconfiguring during the transportation mission process to grasp the object. Still, the structural design of each module limits the application of the whole system to the shape of the target operating object, especially for the thin panel which is hard to hold from a flat plane during the transportation mission. In~\cite{li2023wrench}, a multi-drone system with a self-adaptive gripper is developed for grasping and transportation tasks. In this paper, we choose to install a rigid rod underneath the drone's base with a gripper on its end as shown in Fig.\ref{fig_drone}. This structure design not only introduces the control complexity for the whole dynamics of the vehicle but also increases the detachability for different shapes of the object. Additionally, all aforementioned works limit their drones' structures in the directions of all propellers along $z$-axis in its body frame, which can only generate the wrench $\mathbf{w} =  \left[f_z, \boldsymbol{\tau}\right]^\intercal\in \reals^4$ in four dimensions where $ \boldsymbol{\tau} = \left[\tau_x, \tau_y, \tau_z \right]^\intercal \in \reals^3$. The limited structure lost $f_x$ and $f_y$ generated by the 4 motors of the quadrotor, which would also limit the performance of the transportation mission in 3D space. In this paper, we consider using a fully actuated hexarotor UAV~\cite{flores2022fully,hexa2015franchi} in which each rotor has been tilted to a specific angle along 2 axes to give the drone full actuation in 6 DoF.

\vspace{-0.5pt}
Manipulating and interacting with partially known environments is challenging as it involves adapting to different operational conditions, especially for missions using multiple aerospace robots \cite{ollero2022past,khamseh2018aerial,yang2022collaborative}. One of the main drawbacks of the common methods is their lack of adaptability when the object's parameters to service or transport are unknown \cite{culbertson2021decentralized}. Similarly, robot manipulators performing a contact-based inspection of a variable friction surface would have difficulty maintaining contact with the surface during the inspection task \cite{aghili2019robust}. In this case, we design a novel decentralized adaptive controller in this paper to implement on two different aerospace-manipulated systems: one composed of fully actuated hexarotor UAVs and another one composed of a group of space tugs, in the same transportation tasks under gravity and zero-gravity environment to examine the feasibility of our controller design.

The \textbf{contribution} of this paper is threefold: \textbf{(i)} We combine a grasping system and a fully actuated hexarotor UAV to output desired wrench $\boldsymbol{w} \in \reals^6$ in 6 DoF on grasping system to hold and transport the object, which improves the detachability and actuation performance during our transportation task in 3D space, and \textbf{(ii)} introduce a decentralized adaptive method implemented on multiple aerospace systems that operate effectively under uncertainties, including unknown mass, inertia and the location of grasping points, and \textbf{(iii)} we extend our work on multiple space tugs under a zero-gravity environment by testing the feasibility of our decentralized adaptive controller design. This method represents a robust solution to the longstanding challenges associated with satellite detumbling.
 
% \pagebreak
\section{Decentralized Adaptive Transportation Problem}
\label{prob_stmt}
% \begin{wrapfigure}{r}{0.35\textwidth}
% 	\vspace{-9mm}
% 	\includegraphics[width=\linewidth]{Figs/fig2_v2.pdf}
% 	\caption{Fully actuated hexarotor UAV with a grasping mechanism.
%  % \todo{the gripper and the propeller do not need a frame.}
%  }
% 	\label{fig_drone}
% 	% \vspace{-7mm}
% \end{wrapfigure}
This section describes the transportation problem for a free-floating rigid panel in the shape of a convex polygon through the implementation of our decentralized adaptive controller design, including the dynamics of the two different aerospace systems and the grasping dynamics of each agent to the payload.

\subsection{Robots and Reference Frames}
This paper proposes a novel methodology for transporting a rigid polygon panel using multiple drones and then extends the work by using a group of space tugs in space. 
We consider a team of $n$ vehicles, indexed by $i=1,...,n$, that can be either drones or space tugs. 

\begin{definition}[Drone]
% \textbf{Drone:}
A drone is a fully actuated hexarotor \cite{hexa2015franchi,flores2022fully} with a grasping mechanism composed of a rigid rod and a gripper as shown in Fig.~\ref{fig_drone}. Once the drone grasps the object, the connection is considered rigid.
\end{definition}

It comprises six rotors, each rotor $j$ generates a wrench denoted as $\mathbf{w}_{i_j} = \left[\boldsymbol{f}_{i_j}^\intercal, \boldsymbol{\tau}_{i_j}^\intercal \right]^\intercal$  $\in \reals^6$ where $\boldsymbol{f}_{i_j}=\left[0,0, k_f \, \omega_{i_j}^2 \right]^\intercal$ is the force and $\boldsymbol{\tau}_{i_j}= \left[0,0,k_m \,\omega_{i_j}^2 \right]^\intercal$ is the torque that the rotor generates when it rotates at an angular velocity $\omega_{i_j}$. The coefficients $k_f$ and $k_m$ can be obtained experimentally. 
 The rotors are tilted as pictured in Fig.~\ref{fig_drone}.

\begin{definition}[Space Tug]
% \textbf{Space Tug:}
A space tug is an independently operating, self-propelled spacecraft for assembly, maintenance, repair, and contingency operations in $SE(3)$~\cite{gao2023autonomous} (see~Fig.~\ref{fig:tug}).

\end{definition}

% \begin{wrapfigure}[12]{r}{0.35\textwidth}
% 	\vspace{-10mm}
% 	\includegraphics[width=\linewidth]{Figs/Fig_tug.png}
% 	\caption{Space tug with 6 rockets around its sides. The red arrow shows each rocket proposition in its direction.}
%  % \todo{$f_i$ is not defined}}
% 	\label{fig:tug}
% 	% \vspace{-7mm}
% \end{wrapfigure}
The tug is propelled using six rockets in different directions to push or pull in 3D space, forming a fully actuated vehicle that generates a wrench $\mathbf{w}_i \in \reals^6$.
Each vehicle  $i$, either drone or space tug, has a mass $m_i$ and moment of inertia $\boldsymbol{J}_i$. Note that we will use drones and space tugs to process the same transportation task at different gravity levels to prove the feasibility of our controller design.

\begin{figure}[H]
\centering
\includegraphics[width=1\linewidth]{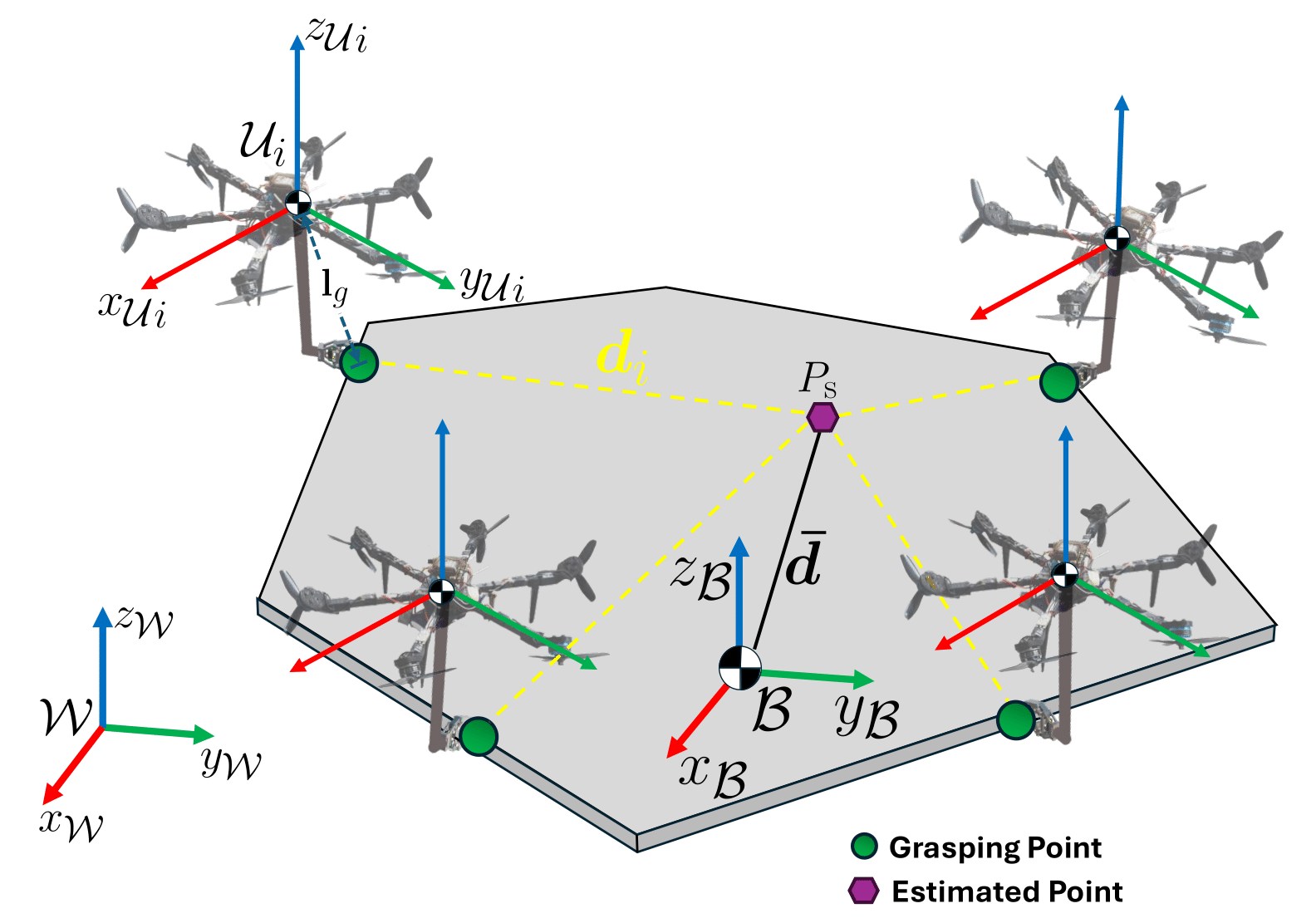}
\caption{A group of fully actuated hexarotor UAVs grasping a free-floating object during the transportation task in 3D space.}
\label{fig_2}
\end{figure}

% \todo{What is the difference between the hexarotor and the tug?}
The coordinate systems are illustrated in Fig.~\ref{fig_2}. The world frame $\mathcal{W}$ is 
fixed and has its $z$-axis pointing upward. We consider $n$ vehicles using grippers to grasp a rigid object. The body frame $\mathcal{B}$ is located on the object's center of mass. It is assumed that the body frame axes are chosen as the principal axes of the entire system. Each vehicle has an individual body frame $\mathcal{U}_i$, attached to its center of mass with $z_{\mathcal{U}_i}$  perpendicular to the plane of the base and pointing up. We require the $z_{\mathcal{U}_i}$ axes and $z_\mathcal{B}$ to be parallel. We use Roll-Pitch-Yaw to model the rotation of frame $\mathcal{B}$ and frame $\mathcal{U}_i$ in the world frame~$\mathcal{W}$. 
We set up one estimated point $P_s$ on the target object that each vehicle can share their common measurement information $\boldsymbol{d}_i$ which is from each grasping point to $P_s$ to estimate the object's unknown properties concerning the unknown information $\boldsymbol{\bar d}$, which is from $P_s$ to the origin of the frame $\mathcal{B}$. 
Let $\boldsymbol{q} = [\boldsymbol{p}^\intercal, \boldsymbol{\theta}^\intercal]^\intercal \in \reals^{6 \times 1}$ be the position and orientation of the manipulated object, where $\boldsymbol{p} \in \reals^3$ and $\boldsymbol{\theta} \in \reals^{3}$ represent the pose of $\mathcal{B}$ with respect to $\mathcal{W}$.  
The rotation matrix $\mathbf{^\mathcal{B}_{\mathcal{W}}R} \in SO(3)$ denotes the orientation of $\mathcal{B}$ with respect to $\mathcal{W}$ and  $\mathbf{^{\mathcal{U}_i}_{\mathcal{W}}R}$ indicates the rotation matrix of $\mathcal{U}_i$ with respect to $\mathcal{W}$.

\subsection{Dynamics}

% The dynamics of the $n$ vehicles attached to the object are modeled using the Euler-Largrian equation,
The dynamics of the free-floating object attached by $n$ vehicles can be modeled using the Euler-Lagrange equation,
\begin{equation}
   \begin{aligned}
    \mathbf{w} = \boldsymbol{M}(\boldsymbol{q})\ddot{\boldsymbol{q}} + \boldsymbol{C}(\boldsymbol{q}, \dot{\boldsymbol{q}})\dot{\boldsymbol{q}} + \boldsymbol{g}, \label{eq:service}
    \end{aligned} 
\end{equation}
where $\mathbf{w} = \begin{bmatrix}
    \boldsymbol{f},\boldsymbol{\tau}
\end{bmatrix}^\intercal \in \reals^{6\times 1}$ is the total wrench  generated by the $n$ vehicles,
where $\boldsymbol{f}$ is the total force, and $\boldsymbol{\tau}$ is the total torque with respect to the point $P_s$ in the frame $\mathcal{B}$. 
%from the $n$ Drones' grippers, which is comprised of both the force $\boldsymbol{f} = \begin{bmatrix}
%     f_x, f_y, f_z
% \end{bmatrix}^\intercal \in \reals^3$ and torque $\boldsymbol{\tau} = \begin{bmatrix}
%     \tau_x, \tau_y, \tau_z
% \end{bmatrix}^\intercal \in \reals^3$ applied on the client at the point $\boldsymbol{\boldsymbol{P_s}}  = \left[x_p, y_p, z_p  \right]^\intercal \in \reals^3$ in frame $\mathcal{B}$. 
The positive semi-definite inertia matrix of the object is
% related to the measurement point $\boldsymbol{P_d}$, where
\begin{equation*}
    \begin{aligned}
    \boldsymbol{M}(\boldsymbol{q}) = \begin{bmatrix}
    m\boldsymbol{\mathrm{I}} & -m(\mathbf{^\mathcal{B}_\mathcal{W}R}\boldsymbol{\bar d})^\times\\
    -m(\mathbf{^\mathcal{B}_\mathcal{W}R}\boldsymbol{\bar d})^\times & \mathbf{^\mathcal{B}_\mathcal{W}R}\boldsymbol{J_d}\mathbf{^\mathcal{B}_\mathcal{W}R^\intercal} 
    \end{bmatrix},
    \end{aligned}
\end{equation*}
% \todo{the operator ${}^\times$ is not defined.}
% is the inertia matrix $\left(S.P.D\right)$ of the service module. 
% \todo{change the moment of inertia to $\boldsymbol{J}$ since $\boldsymbol{I}$ is being used for the identity matrix}
where $\boldsymbol{\mathrm{I}} \in \mathbb{R}^{3 \times 3}$ is the identity matrix, $\boldsymbol{\bar d} \in \reals^3$ and $\boldsymbol{J}_d$ are, respectively, the position vector of the estimated point concerning the center of mass and moment of inertia of the client in frame $\mathcal{B}$. $(\cdot)^\times$ denotes the map from $\reals^3$ to $SO(3)$,  which is skew-symmetric.
The moment of inertia is obtained using the parallel axis theorem~\cite{khalil2002nonlinear},
% $\boldsymbol{d}_i$ denoted the position of measurement point $\boldsymbol{P}_d$ the position of $\it{i}$-th space tugs in frame $\mathcal{B}$, respectively, 
% $\boldsymbol{I}_d$ is the moment of interia in our measurement point based on the Parallel axis theorem\cite{khalil2002nonlinear}
\begin{equation*}
  \begin{aligned}
    \boldsymbol{J_d} =  \boldsymbol{J_{cm}} + \boldsymbol{m}((\boldsymbol{\bar d^\intercal \bar d})\boldsymbol{\mathrm{I}} - \boldsymbol{\bar d \bar d^\intercal} ), 
\end{aligned}  
\end{equation*}
where $\boldsymbol{J_{cm}}$ is the moment of inertia about the center of mass. The centripetal and Coriolis matrix is

\begin{equation*}
\resizebox{0.92\hsize}{!}{$
    \boldsymbol{C}(\boldsymbol{q}, \boldsymbol{\dot{q}})=\begin{bmatrix}
        \mathbf{0}_{3 \times 3} & -m \boldsymbol{\omega}^{\times}\left(\mathbf{\mathbf{^\mathcal{B}_\mathcal{W}R} \bar d}\right)^{\times} \\
-m \boldsymbol{\omega}^{\times}\left(\mathbf{\mathbf{^\mathcal{B}_\mathcal{W}R} \bar d}\right)^{\times} & \boldsymbol{\omega}^{\times}\mathbf{^\mathcal{B}_\mathcal{W}R} \boldsymbol{J}_d\mathbf{^\mathcal{B}_\mathcal{W}R}-m\left(\left(\mathbf{^\mathcal{B}_\mathcal{W}R} \mathbf{\bar d}\right)^{\times} \dot{\mathbf{q}}\right)^{\times}
    \end{bmatrix}$}.
\end{equation*}

The term $\boldsymbol{g}$ is a $6 \times 1$ vector representing the gravitational wrench that only applies to the scenario of aerial vehicles.

\subsection{Vehicle's Wrench}
\label{dyns_uav}
The rotors in the vehicle generate a wrench $\mathbf{w}_{p_i} = \left[\boldsymbol{f}_{p_i}, \boldsymbol{\tau}_{p_i} \right]^\intercal $ with respect to the center of mass of the $i^{\text{th}}$ vehicle in frame $\mathcal{U}_i$, in which $\boldsymbol{f}_{p_i}$ and $ \boldsymbol{\tau}_{p_i}$ are defined as
% \begin{equation}
% \boldsymbol{f}_{p_i}  = \sum_{j=0}^{6} \mathbf{^{\mathcal{M}_{i_j}}_{\mathcal{U}_i}R}  \boldsymbol{f}_{i_j},
% \label{Force_propeller}
% \end{equation}

% \begin{equation}
% \boldsymbol{\tau}_{p_i}  = \sum_{j=0}^{6} \boldsymbol{d}_{i_j} \times \mathbf{^{\mathcal{M}_{i_j}}_{\mathcal{U}_i}R} \boldsymbol{f}_{i_j}  + \sum_{j=0}^{6} \mathbf{^{\mathcal{M}_{i_j}}_{\mathcal{U}_i}R}  \boldsymbol{\tau}_{i_j},
% \label{Torque_propeller}
% \end{equation}
 \begin{equation}
     \begin{aligned}
         \boldsymbol{f}_{p_i} &= \sum_{j=0}^{6} \mathbf{^{\mathcal{M}_{i_j}}_{\mathcal{U}_i}R}  \boldsymbol{f}_{i_j}\\
         \boldsymbol{\tau}_{p_i} &= \sum_{j=0}^{6} \left(\mathbf{^{\mathcal{M}_{i_j}}_{\mathcal{U}_i}R}\boldsymbol{f}_{i_j} \right)^\times \boldsymbol{d}_{i_j}    + \sum_{j=0}^{6} \mathbf{^{\mathcal{M}_{i_j}}_{\mathcal{U}_i}R}  \boldsymbol{\tau}_{i_j},
     \end{aligned}
 \end{equation}
where $\boldsymbol{d}_{i_j} \in \reals^3 $ denotes the position of the $j^{\text{th}}$ motor in frame $\mathcal{U}_i$ and the rotation matrix
$^{\mathcal{M}_{i_j}}_{\mathcal{U}_i}\mathbf{R} \in SO(3)$ defines the rotor's orientation with respect to $\mathcal{U}_i$.
It is assumed that the rotors are in different orientations to form a fully actuated vehicle following the conditions in \cite{hexa2015franchi}.

The wrench $\mathbf{w}_i = \left[\boldsymbol{f}_i, \boldsymbol{\tau}_i \right]$ generated by $i^{\text{th}}$ vehicle in frame $\mathcal{W}$ can be calculated using the Newton-Euler's equation,
% \todo{I am not sure this equation is correct}

\begin{equation}
    \begin{aligned}
        \boldsymbol{f}_i &= \mathbf{^{\mathcal{U}_i}_{\mathcal{W}}R}\boldsymbol{f}_{p_i}  -m_i\dot{\boldsymbol{v}}_i + \boldsymbol{g}_i\\
        \boldsymbol{\tau}_{i} &= \mathbf{^{\mathcal{U}_i}_{\mathcal{W}}R}(- \left(\boldsymbol{J}_i \boldsymbol{\omega}_i \right)^\times \boldsymbol{\omega}_i + \boldsymbol{\tau}_{p_i} - \boldsymbol{J}_i \dot{\boldsymbol{\omega}}_i) ,
    \end{aligned}
    \label{eq2_uav_dynamics}
\end{equation}
where $\boldsymbol{v}_{i}\in \mathbb{R}^{3}$ denote the 
linear velocity of the  $i^{\text{th}}$ vehicle in frame $\mathcal{W}$, $\boldsymbol{\omega}_i \in \mathbb{R}^{3}$ represent the angular velocity of the  $i^{\text{th}}$ vehicle in frame $\mathcal{U}_i$, and $\boldsymbol{g}_i$ is the vector of the gravitational force of the $i^{\text{th}}$ vehicle that is zero in the space environment.

\subsection{Grasping Dynamics of Manipulator System}
\label{grasp_dynamics}

Next, we model the wrench $\mathbf{w}$ applied to the target object by an aerospace robot. 
The total wrench  applied on the object is the sum of the wrenches $\mathbf{w}_i$ applied by each robot
\begin{equation}
    \begin{aligned}
\mathbf{w}=\sum_{i=1}^n \boldsymbol{G}\left(\boldsymbol{q}, \boldsymbol{d}_i\right) \mathbf{w}_i, \label{eq6}
\end{aligned}
\end{equation}
where the matrix $\boldsymbol{G}$ 
% \todo{Matrix notation should be  $\boldsymbol{G}$}
maps the forces and torques produced by each robot into wrenches on the object 

% \begin{equation}
% \begin{aligned}
%  \boldsymbol{p}_e &= \boldsymbol{p}_{\mathcal{U}_i}+ \mathbf{^{\mathcal{U}_i}_{\mathcal{W}}R}\boldsymbol{l}_{g}, \\
%  \mathbf{^{e}_{\mathcal{W}}R} &= \mathbf{^{\mathcal{U}_i}_{\mathcal{W}}R} \mathbf{^e_{\mathcal{U}_i}R},
% \end{aligned}
% \end{equation}

\begin{equation*}
    \begin{aligned}
    \boldsymbol{G}\left(\boldsymbol{q}, \boldsymbol{d}_i\right) = \begin{bmatrix}
    \boldsymbol{\mathrm{I}} & \mathbf{0}_{3 \times 3}\\
    (\mathbf{^\mathcal{B}_\mathcal{W}R}\, \mathbf{d}_i)^{\times}(\mathbf{^{\mathcal{E}_i}_\mathcal{W}R}\,\mathbf{l}_g)^{\times} & \boldsymbol{\mathrm{I}}
\end{bmatrix}.
\end{aligned}
\end{equation*}
where $\mathbf{^{\mathcal{E}_i}_{\mathcal{W}}R} = \mathbf{^{\mathcal{U}_i}_{\mathcal{W}}R} \mathbf{^{\mathcal{E}_i}_{\mathcal{U}_i}R}$ maps the transforming coordinates from frame $\mathcal{E}_i$ to frame~$\mathcal{W}$ and $\mathbf{l}_g \in \reals^3$ is the position of the origin of frame $\mathcal{U}_i$ relative to grasping position in frame $\mathcal{B}$. Combined with the dynamics in~\eqref{eq:service}, we obtain
% We then take the equation and then we can get
\begin{equation}
    \begin{aligned}
    \boldsymbol{M}(\boldsymbol{q}) \ddot{\boldsymbol{q}}+\boldsymbol{C}(\boldsymbol{q}, \boldsymbol{\dot{q}}) + \boldsymbol{g} = \sum_{i=1}^n \boldsymbol{G}(\boldsymbol{q}, \boldsymbol{d_i}) \mathbf{w}_i. \label{eq7}
\end{aligned}
\end{equation}

Our objective is to design a decentralized control law for the wrench $\mathbf{w}_i$ from each robot to the client that provides reference tracking despite parametric uncertainty about the dynamics of the client, $\boldsymbol{d}_i$ and $\boldsymbol{\bar d}$. 

\subsection{Decentralized Adaptive Transportation}
\label{DAC_T}
In this section, we will design a decentralized adaptive controller for our transportation task to let the group of vehicles manipulate the object to track the desired position trajectory. First, 
we introduce a composite error $\boldsymbol{s}$: 
\begin{equation}
    \begin{aligned}
    \boldsymbol{s} = \begin{bmatrix}
    \boldsymbol{\epsilon} \\
    \boldsymbol{o}
\end{bmatrix} = \begin{bmatrix}
    \dot{\boldsymbol{e}}_p  + \beta \boldsymbol{e}_p\\
    \boldsymbol{e}_w +  \dfrac{1}{2}\beta\mathbf{^\mathcal{B}_\mathcal{W}R}_d(\mathbf{^\mathcal{B}_\mathcal{W}R}_e + \mathbf{^\mathcal{B}_\mathcal{W}R}_e^\intercal) %\label{eq:error}
\end{bmatrix} \label{eq:error}
\end{aligned},
\end{equation}
where $\boldsymbol{e}_p(t) = \boldsymbol{p}_d(t) - \boldsymbol{p}(t)$ is the position tracking error, $\beta < 1$ is a constant positive number, $\boldsymbol{\dot{e}}_p(t)$ is the velocity tracking error, $\boldsymbol{e}_w(t)$ is the angular velocity error, $\mathbf{^\mathcal{B}_\mathcal{W}R}_d$ is the desired rotation matrix of $\mathcal{B}$ with respect to $\mathcal{W}$, and $\mathbf{^\mathcal{B}_\mathcal{W}R}_e = \mathbf{^\mathcal{B}_\mathcal{W}R}_d^\intercal\mathbf{^\mathcal{B}_\mathcal{W}R}$. Considering the rotation of the target object during the transportation tasks, we use the ‘‘reference rotation error" $\bold{Rot_{err}} = tr(\mathbf{^\mathcal{B}_\mathcal{W}R}_e - \boldsymbol{\mathrm{I}})$ to denote the rotation error. We need $\bold{Rot_{err}} \rightarrow 0 $ and $\boldsymbol{e}_p, \boldsymbol{e}_w \rightarrow 0$ as $t \rightarrow \infty$.

The controller employs the regressor matrix    $\boldsymbol{Y}_\varphi(\boldsymbol{q}, \boldsymbol{\dot{q}}, \boldsymbol{\dot{q}}_d, \boldsymbol{\ddot{q}}_d)$,
\begin{align}
    \sum_{i=1}^n \boldsymbol{Y}_{\varphi}\boldsymbol{\widetilde{\varphi}}_i = \sum_{i=1}^n \gamma(\boldsymbol{\widetilde{M}} \ddot{\boldsymbol{q}}_d + \boldsymbol{\widetilde{C}\dot{q}}_d + \boldsymbol{\widetilde g}), \label{eq:regressor-phi}
\end{align}
where $\gamma=\tfrac{1}{n}$ divides the control workload evenly among the agents, $\boldsymbol{\widetilde{\varphi}}_i = \boldsymbol{\hat{\varphi}}_i - \boldsymbol{\varphi}_i$ denotes the error of parameters estimation related to the physical properties of the service module scaled by $\gamma$, $\boldsymbol{\widetilde{M}(u)} =  \boldsymbol{\hat{M}(q)} -  \boldsymbol{M(q)}$ is the inertia error of the stiffness system, $\boldsymbol{\widetilde{C}(\dot{q}, q)} =  \boldsymbol{\hat{C}(\dot{q}, q)} -  \boldsymbol{C(\dot{q}, q)}$ is the Coriolis and centripetal error, and $\boldsymbol{\widetilde{g}} =  \boldsymbol{\hat{g}} -  \boldsymbol{g}$ is the gravity error.

To provide asymptotic tracking despite parametric uncertainty, we use the proportional-derivative control law with adaptive feedback linearization
% with the term, $-\boldsymbol{K_D}s$
\begin{align}
    \mathbf{\hat{w}}_i = \boldsymbol{\hat{G}}_i\mathbf{w}_i = \boldsymbol{Y}_{\varphi} \boldsymbol{\hat{\varphi}}_i - \boldsymbol{K}_{\scriptstyle\text{PD}}\boldsymbol{s} , \label{eq16}
\end{align}
where the term $\boldsymbol{Y}_{\varphi} \boldsymbol{\hat{\varphi}}_i$ feedback linearizes the service satellite dynamics~\eqref{eq:service} using the estimated parameters $ \boldsymbol{\hat{\varphi}}_i$
and the term $-\boldsymbol{K}_{\scriptstyle\text{PD}}\boldsymbol{s}$ provides proportional-derivative control. 
Note that the composite error~\eqref{eq:error} contains both position and orientation errors as well as velocity and angular velocity errors, producing a proportional-derivative controller. 

Now let's introduce another regressor $\boldsymbol{Y}_d(\mathbf{{\hat{w}}}_i, \boldsymbol{q})$ and then we can get
\begin{align}
    \boldsymbol{Y}_d(\mathbf{{\hat{w}}}_i, \boldsymbol{q})\boldsymbol{\hat{d}}_i = -\boldsymbol{\widetilde{G
}}_i \mathbf{{\hat{w}}}_i\;. \label{eq:regressor-g}
\end{align}

Thus, we can get the adaptation laws as
\begin{subequations}
\label{eq:adaptation}
\begin{align}
        \dot{\hat{\boldsymbol{\varphi}}}_i & =-\Gamma_{\varphi} \boldsymbol{Y}_{\varphi}\left(\boldsymbol{q}, \dot{\boldsymbol{q}}, \dot{\boldsymbol{q}}_d, \ddot{\boldsymbol{q}}_d\right)^\intercal\boldsymbol{s}, \\
        \dot{\hat{\boldsymbol{d}}}_i & =-\Gamma_d \boldsymbol{Y}_d(\mathbf{{\hat{w}}}_i, \boldsymbol{q})^\intercal\boldsymbol{s}. \label{eq18}
\end{align}
\end{subequations}

We use a Lyapunov-like function to prove our controller design makes the whole system exponentially stable. Consider the Lyapunov-like function 
\begin{equation}
    \begin{aligned}
    V(t)=\frac{1}{2}\left[\boldsymbol{s}^\intercal \boldsymbol{M} \boldsymbol{s}+\sum_{i=1}^n \widetilde{\boldsymbol{\varphi}}_i^\intercal \boldsymbol{\Gamma} \widetilde{\boldsymbol{\varphi}}_i+\widetilde{\boldsymbol{d}}_i^\intercal \boldsymbol{K} \widetilde{\boldsymbol{d}}_i\right], \label{eq9}
\end{aligned}
\end{equation}
where $\widetilde{\boldsymbol{d}}_i$ is the position error that is the difference between the actual position of the $i$-th tug $\boldsymbol{d}_i$ concerning the measurement point $\boldsymbol{P}_s$ and the tug's estimated $\boldsymbol{\hat{d}}_i$. $\boldsymbol{\Gamma}$ and $\boldsymbol{K}$ are symmetric positive definite matrices related to the adaptive gain, usually diagonal.\\

Then, by taking the derivative, we obtain
\begin{small}
\begin{equation}
\begin{aligned}
\dot V(t) &=\boldsymbol{s}^\intercal(\sum_{i=1}^n \boldsymbol{G}(\boldsymbol{q,d_i}) \mathbf{{w}}_i - \boldsymbol{M} \ddot{\boldsymbol{q}}_d - \boldsymbol{C\dot{q}}_d - \boldsymbol{g})\\
    &+ \sum_{i=1}^n \left[\widetilde{\boldsymbol{\varphi}}_i^\intercal \boldsymbol{\Gamma} \dot{\hat{\boldsymbol{\varphi}}}+\widetilde{\boldsymbol{d}}_i^\intercal \boldsymbol{K} \dot{\hat{\boldsymbol{d}}}_i \right], \label{eq10}
\end{aligned}
\end{equation}
\end{small}
where $\dot{\hat{\boldsymbol{\varphi}}}_i = \dot{\widetilde{\boldsymbol{\varphi}}}_i$ and $\dot{\hat{\boldsymbol{d}}}_i = \dot{\widetilde{\boldsymbol{d}}}_i$ since the parameter $\boldsymbol{\varphi}_i$ and $\boldsymbol{d}_i$ are constant. Then we use the properties of skew-symmetry \cite{slotine1987adaptive} to eliminate the term $ \boldsymbol{s}^\intercal (\dfrac{1}{2}\dot{\boldsymbol{M}} - \boldsymbol{C}) \boldsymbol{s} = 0$ and $\boldsymbol{\dot{q}}_d = \boldsymbol{s} - \boldsymbol{\dot{q}}$. Next,
let's define the control-law as
\begin{align}
    \mathbf{w} = \boldsymbol{\hat{M}} \ddot{\boldsymbol{q}}_d + \boldsymbol{\hat{C}\dot{q}}_d + \boldsymbol{\hat g}\;. \label{eq11}
\end{align}

We substitute \eqref{eq11} to \eqref{eq10}, rewrite $\dot{V}(t)$ using the regressors~\eqref{eq:regressor-phi} and \eqref{eq:regressor-g},
 and then substitute adaptation-laws~\eqref{eq:adaptation} to obtain
% where the parameter estimates are updated using the adaptation-laws~\eqref{eq:adaptation}
    \begin{align}
    \begin{split}
        \dot{V}(t) &= \sum_{i=1}^n [-\boldsymbol{s}^\intercal\boldsymbol{K}_{{\scriptstyle\text{PD}}}\boldsymbol{s} + \boldsymbol{\widetilde{\varphi}}_i^\intercal(\boldsymbol{Y_{\varphi}\widetilde{\varphi}_i} \\
        &+ \boldsymbol{\Gamma}_{\varphi}\dot{\hat{\boldsymbol{\varphi}}}_i) + \boldsymbol{\widetilde d}^\intercal(\boldsymbol{Y_d \widetilde d_i} + \boldsymbol{\Gamma}_d \dot{\hat{\boldsymbol{d}}}_i)]\\
        &= n\left(-\boldsymbol{s}^\intercal \boldsymbol{K}_{\scriptstyle\text{PD}} \boldsymbol{s}\right) \leq 0.\label{eq19}      
    \end{split}
    \end{align}
% \begin{align}
%     \dot{V}(t) = n\left(-\boldsymbol{s}^T \boldsymbol{K}_D \boldsymbol{s}\right) \leq 0. \label{eq20}
% \end{align}

Thus, we get $V(t)>0$ and $\dot{V}(t) \leq 0$. So, the system is stable in the sense of Lyapunov.

\vspace{0.385cm}
\section{Simulation Results}
\label{exp_results}

This section presents two simulation results for the transportation task of the payload in $SE(3)$ using two different aerospace robots where $n=4$ fully actuated hexarotor vehicles with grasping systems in Section~\ref{sec::sim-1} and $n=4$ space tugs in Section~\ref{sec::sim-2}. The simulations are implemented on the MuJoCo~\cite{todorov2012mujoco} platform as shown in Fig. \ref{mujoco_a} and Fig. \ref{mujoco_b} with the object parameters summarized in Table~\ref{tab:task}. The first one employs multiple fully actuated hexarotor UAVs in a gravitational environment, while the second one uses a group of space tugs in a zero-gravity environment.

\begin{table}[H]
\centering
\vspace{0.25cm}
 \begin{tabular}{ccc}
 \toprule 
  Definition & Parameter & Value\\
 \midrule
  Object & $m_b$ & 5 kg\\
  & $I_{x x}$ & $1.4255$ kg$\cdot$m$^2$\\
  & $I_{y y}$ & $1.4255$ kg$\cdot$m$^2$\\
  & $I_{z z}$ & $0.8411$ kg$\cdot$m$^2$\\
  Position & $\mathbf{l}_g$ & $[0.1,0.0,-0.3]$ m\\
  & $\mathbf{\bar d}$ & $[0.74,0.0.1,-0.2]$ m\\
  & $\mathbf{d}_1$ & $[-0.8,1.2,0.1]$ m\\
  & $\mathbf{d}_2$ & $[1,1,0.1]$ m\\
  & $\mathbf{d}_3$ & $[1,-0.7,0.1]$ m\\
  & $\mathbf{d}_4$ & $[-0.7,-1.1,0.1]$ m\\
 \bottomrule
 \end{tabular}
 \caption {Simulation parameters for transportation task.}
 \label{tab:task}
\end{table}

Assume all aerospace robots in both simulations are distributed around the object. Additionally, each robot uses its grasping mechanism to firmly grasp the edge to ensure the module can be transported in any direction and orientation in 3D space. For the desired position trajectory $\boldsymbol{p}_d = \left[p_{d_x}, p_{d_y}, p_{d_z}\right]^\intercal$, we take $\boldsymbol{p}_x = \sin{(\omega_x t)}$, $\boldsymbol{p}_y = \cos{(\omega_yt)}$, and $\boldsymbol{p}_z = 0$. Here we adopt the iterative method in \cite{stoer1996introduction} to simulate the trajectory of the robot $\boldsymbol{p}(t) = \boldsymbol{p}(0) + \dot{\boldsymbol{p}}(t) \cdot t$ in which $\boldsymbol{p}(0)$ is the initial state of the service module. For the rotational trajectory, we assume $\boldsymbol{q}_d \in \reals^4$ in both simulations are $\boldsymbol{q}_d = [1, 0, 0, 0]^\intercal$, meaning the object should be kept in a fixed orientation during the transportation task.
% define the quaternion term $\boldsymbol{q} = [w, q_1, q_2, q_3]$ in our Python code and using the exponential map in \cite{kuipers1999quaternions} to transfer quaternion term to SO(3) form as $\boldsymbol{Rot}$ and using the method in \cite{stoer1996introduction} to get the rotation trajectory as 
% \begin{equation}
%     \boldsymbol{Rot}_{t+1} = \exp(\delta \boldsymbol{\omega}^{\times})\boldsymbol{Rot}_t
%     \label{rot_it}.
% \end{equation}
% We will also use this term in both two cases in Task-2.

\subsection{Simulation-1: Transportation Task in Earth}
\label{sec::sim-1}

%\begin{figure}[!h]
% \vspace{-17cm}
%\centering
%\captionsetup[sub]{font=scriptsize,labelfont={sf},oneside,margin={0.75cm,0cm}}
%\begin{subfigure}{2.30in}
 %\includegraphics[width=\textwidth]{Figs/mujoco_drone_fig0_v4.png}
 %\caption{}
 %\label{mujoco_a}
%\end{subfigure}
%\hfill
%\begin{subfigure}{2.38in}
% \includegraphics[width=\textwidth]{Figs/mujoco_tpod_fig0_v2.png}
 %\caption{}
 %\label{mujoco_b}
%\end{subfigure}
%\caption{Transportation task implemented on different aerospace robots on MuJoCo platform. (a) shows the task by multiple fully actuated hexarotor UAVs with a grasping system in the Earth (with gravity); (b) shows the task by a group of space tugs in space (without gravity).}
%\label{fig_mujoco}
%\end{figure}

\begin{figure}[H]
  \centering
  \captionsetup[sub]{font=scriptsize,labelfont={sf},oneside}
  \begin{subfigure}{.5\textwidth}
  \centering
  \includegraphics[width=0.95\linewidth]
  {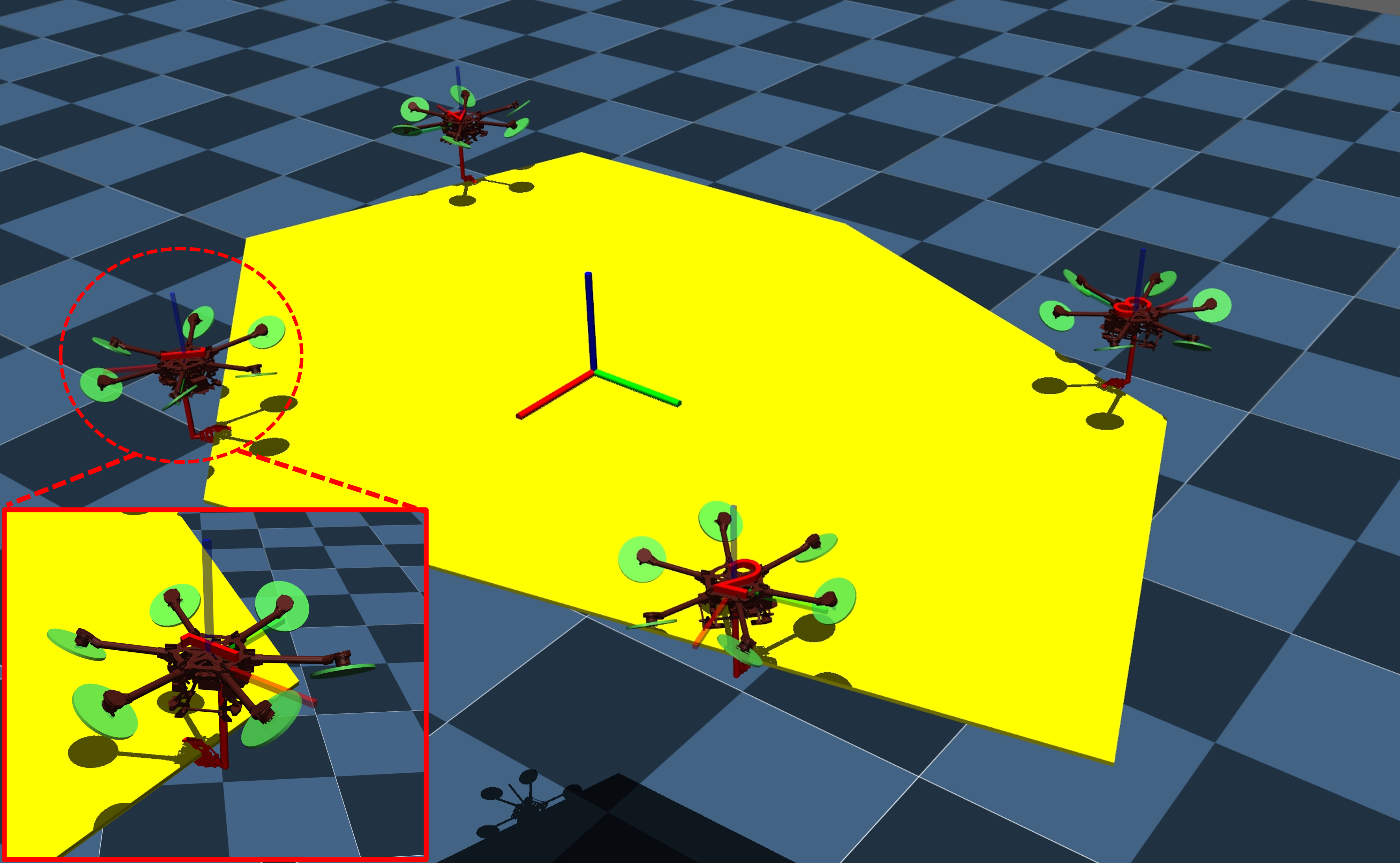}
  \vspace{0.6cm}
  \caption{ }
  \label{mujoco_a}
\end{subfigure}%
\begin{subfigure}{.5\textwidth}
  \centering
  \includegraphics[width=0.97\linewidth]{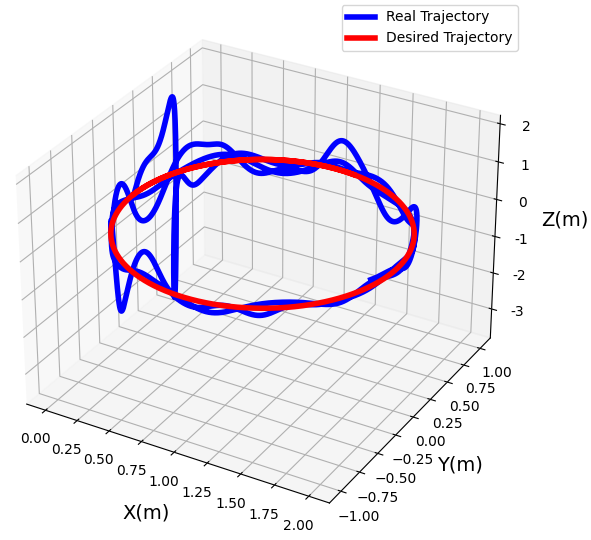}
   \caption{ }
  \label{sim1-a}
\end{subfigure}\\
\vspace{0.2cm}
  \begin{subfigure}{.5\textwidth}
  \centering
  \includegraphics[width=0.95\linewidth]{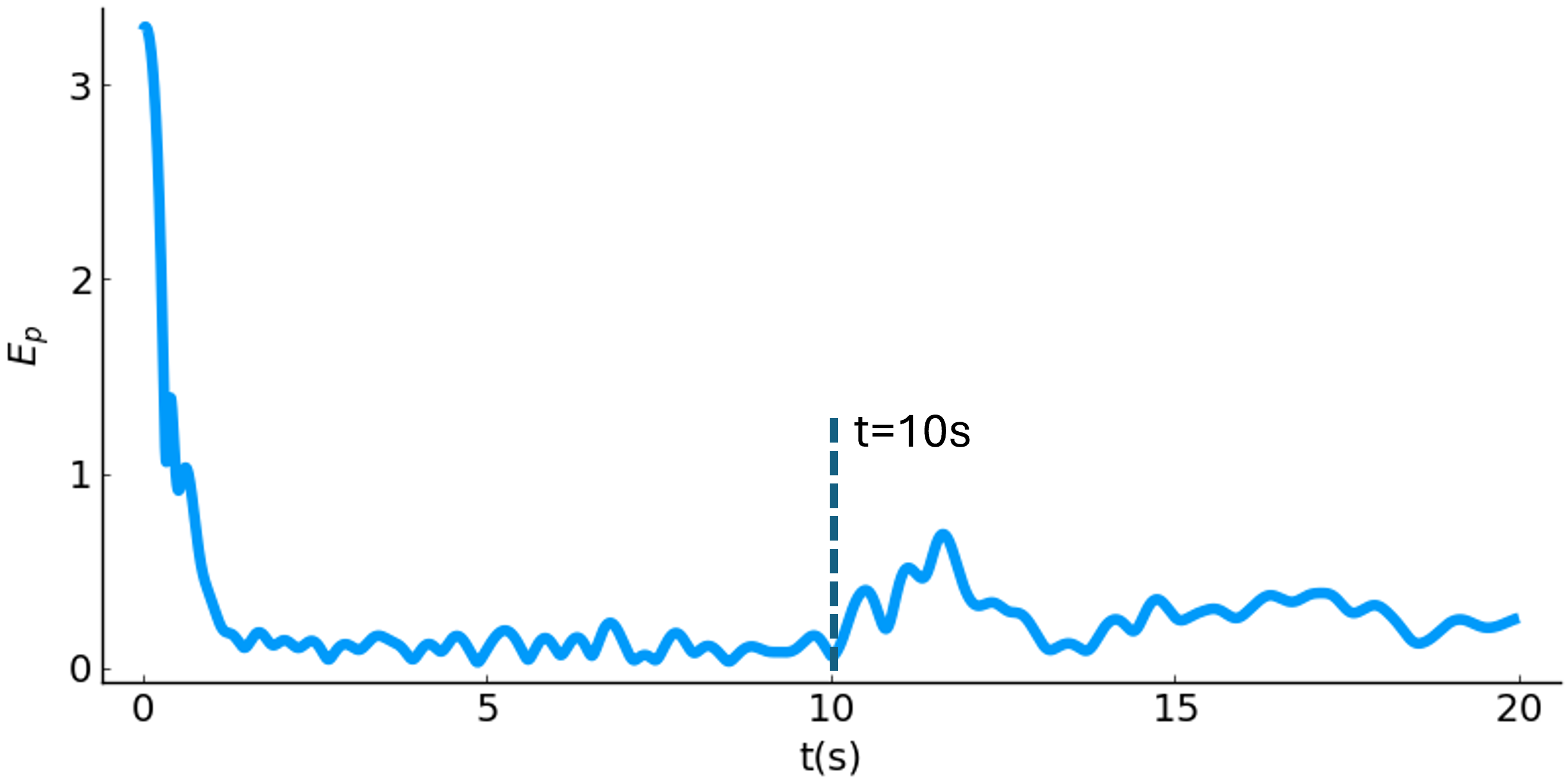}
  \caption{ }
  \label{sim1-b}
\end{subfigure}%
\begin{subfigure}{.5\textwidth}
  \centering
  \includegraphics[width=0.95\linewidth]{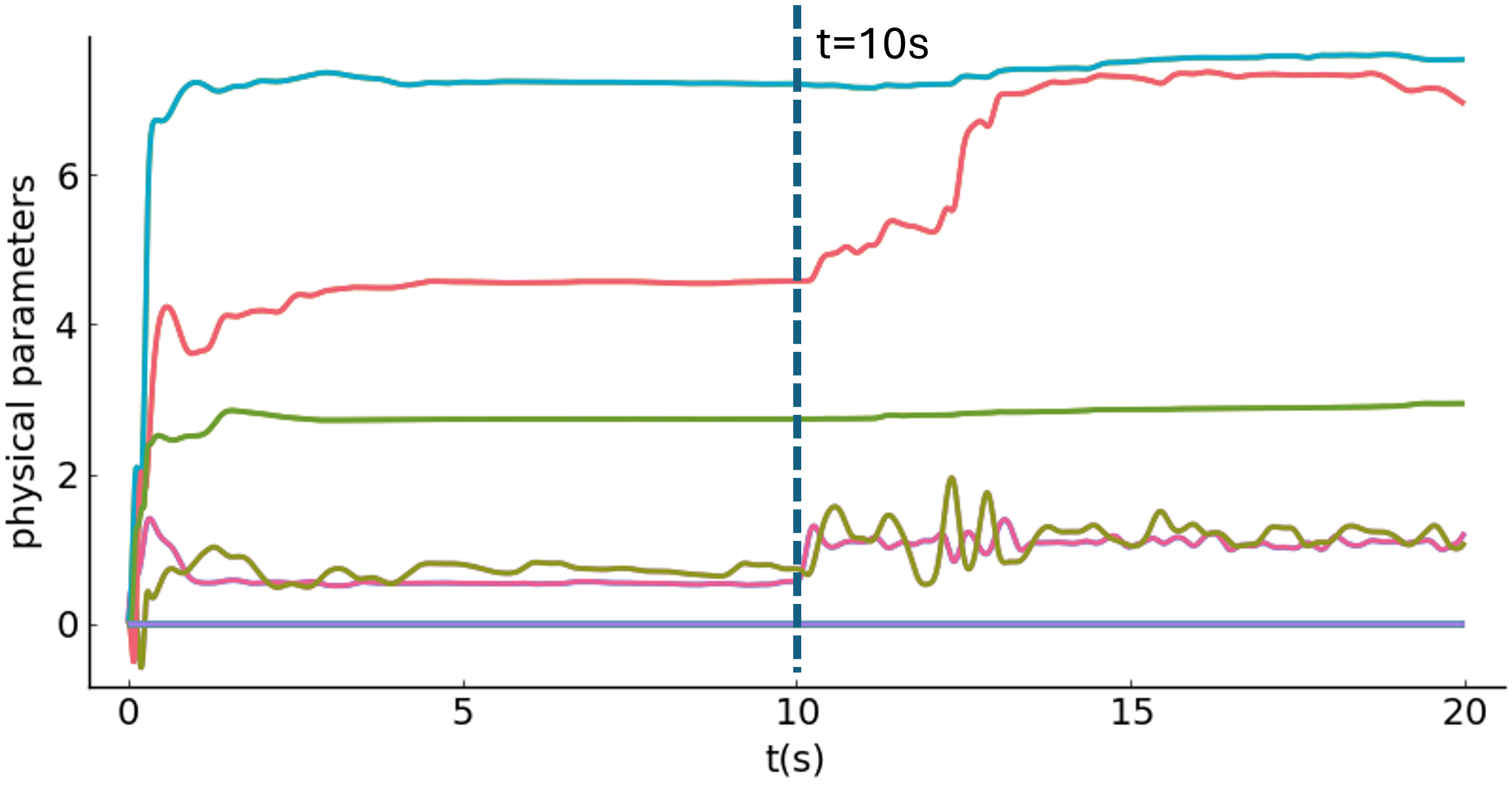}
   \caption{ }
  \label{sim1-c}
\end{subfigure}\\
\vspace{0.2cm}
\begin{subfigure}{.5\textwidth}
  \centering
  \includegraphics[width=0.95\linewidth]{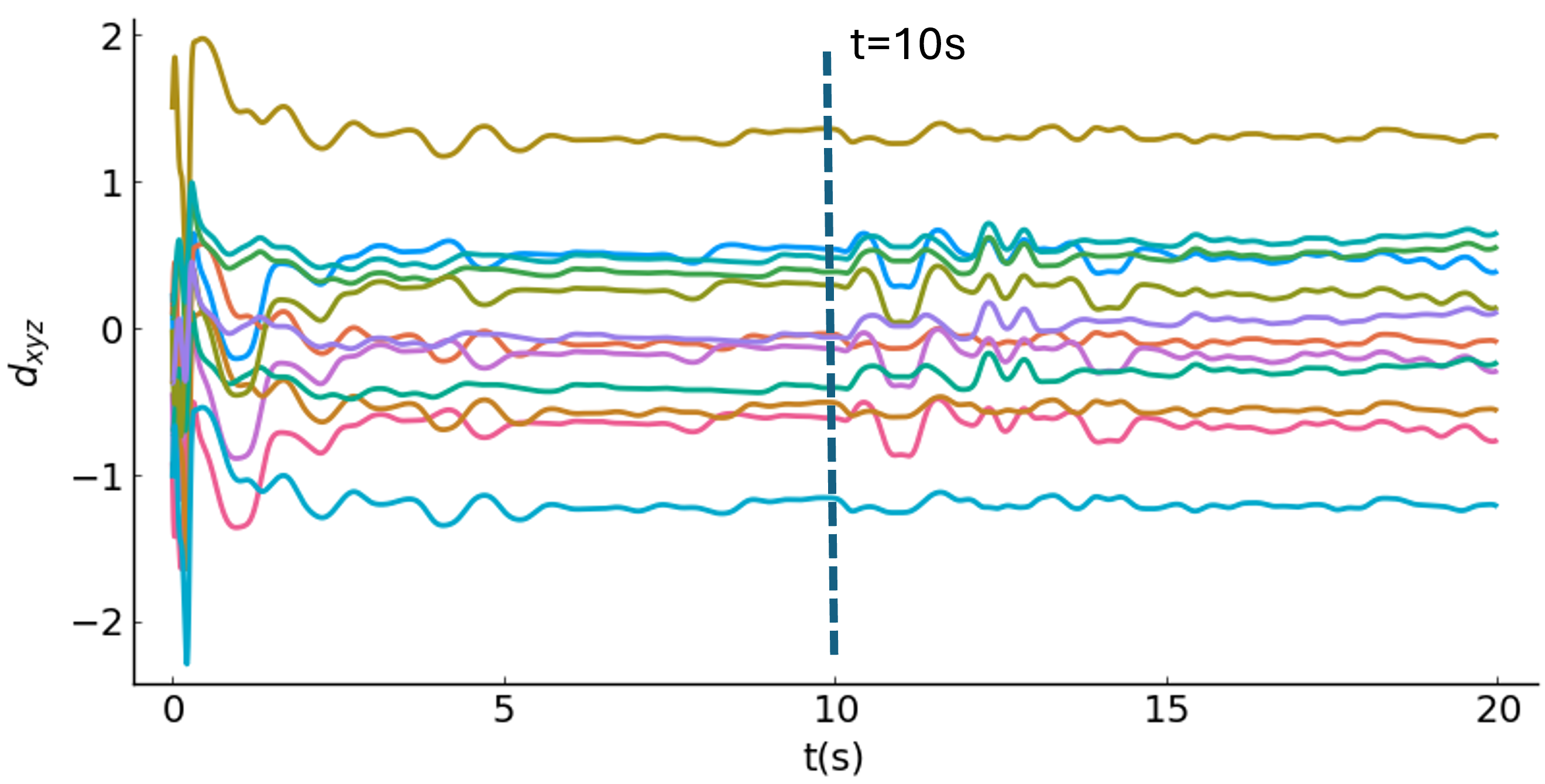}
  \caption{ }
  \label{sim1-d}
\end{subfigure}%
\begin{subfigure}{.5\textwidth}
  \centering
  \includegraphics[width=0.95\linewidth]{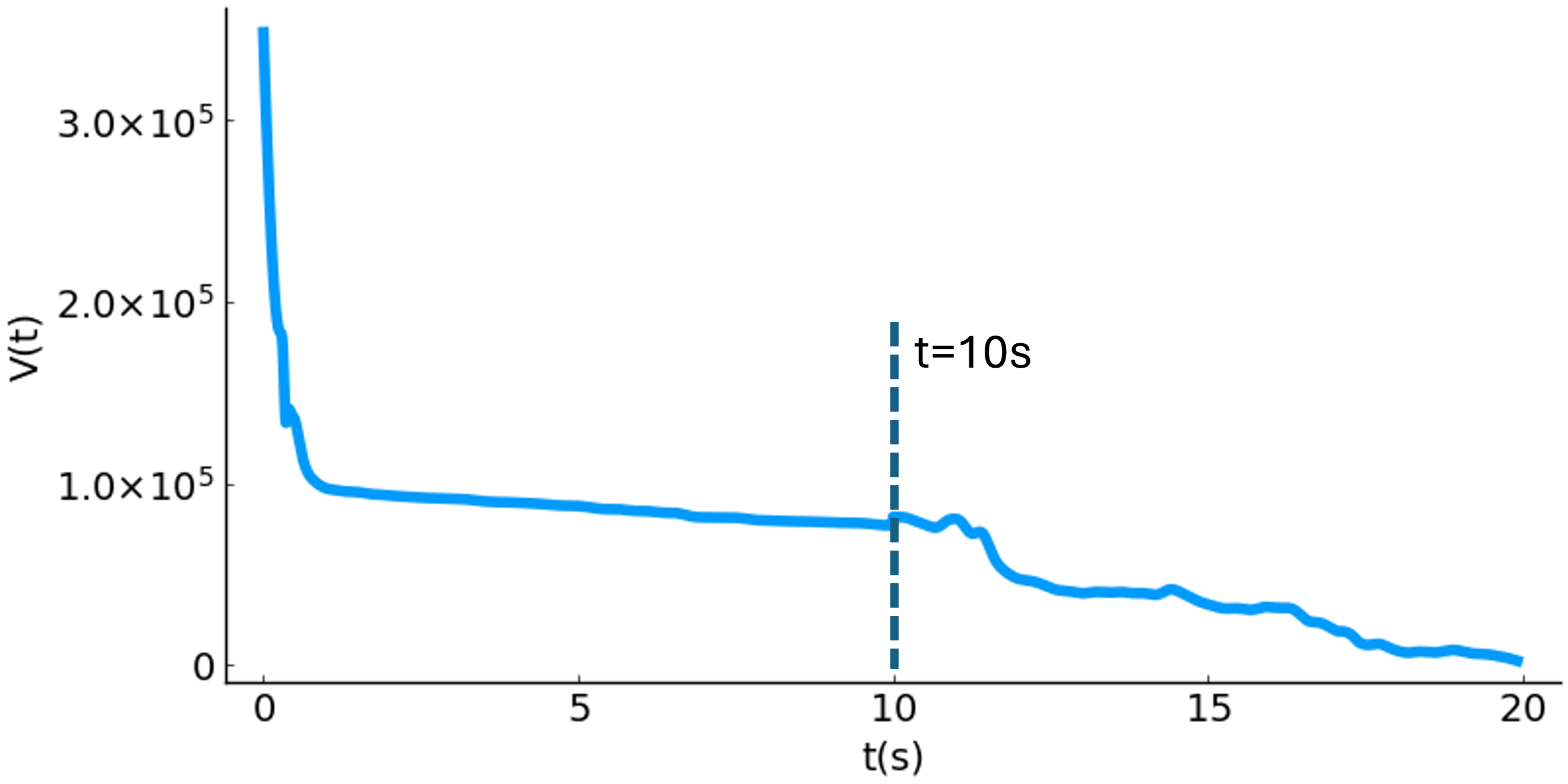}
   \caption{ }
  \label{sim1-e}
\end{subfigure}
\caption{(a) Transportation task implemented on multiple fully actuated hexarotor UAVs with a grasping system in Earth on MuJoCo platform. (b)-(f) Simulation results for using $n$ UAVs collaboratively to transport the payload in $SE(3)$ under gravity environment.}
\end{figure}

Fig. \ref{sim1-a} shows the tracking effect over time in 3D space. Fig. \ref{sim1-b} shows the composite error~\eqref{eq:error} for the client manipulated by a group of hexarotors in the Earth environment. The results show at $10$ s, we disable UAV-1, which introduces a disturbance to the system.

Fig. \ref{sim1-c} and Fig. \ref{sim1-d} show the parameter estimation for the physical properties of the client and the contact point for each aerial vehicle to the measurement point $\boldsymbol{P}_s$. Based on the results, we can see that all signals are bounded, and the final values are close to the actual values.

Fig. \ref{sim1-e} shows the Lyapunov-like function $V(t)$ versus time for the simulation results. The Lyapunov-like function is positive $V(t) > 0$ and $\dot{V}(t) \leq 0$ as time increases. This verifies that our system is stable in the sense of Lyapunov.

%\begin{figure}[H]
%  \centering
%  \captionsetup[sub]{font=scriptsize,labelfont={sf},oneside,margin={0.75cm,0cm}}
%  \subfloat[]{\includegraphics[width=2.in]{Results/3D_UAV.png} \label{sim1-a}}\\
 % \qquad 
 % \subfloat[]{\includegraphics[width=2.in]{Results/UAV_pos_error_v1.png}  \label{sim1-b}} 
 % \qquad 
 % \subfloat[]{\includegraphics[width=2.in]{Results/UAV_physic_est_v1.png}  \label{sim1-c}} 
 % \qquad 
 % \subfloat[]{\includegraphics[width=2.in]{Results/UAV_distance_est_v1.png} \label{sim1-d} }  
 % \qquad 
  %\subfloat[]{\includegraphics[width=2.in]{Results/UAV_lya_v1.png} \label{sim1-e}}
  %\caption{Simulation results for using $n$ UAVs collaboratively to transport the payload in $SE(3)$ under gravity environment.}
 % \label{fig:sim-1}
%\end{figure}

\subsection{Simulation-2: Transportation Task in Space}
\label{sec::sim-2}
\begin{figure}[H]
  \centering
  \captionsetup[sub]{font=scriptsize,labelfont={sf},oneside}
  \begin{subfigure}{.5\textwidth}
  \centering
  \includegraphics[width=0.95\linewidth]{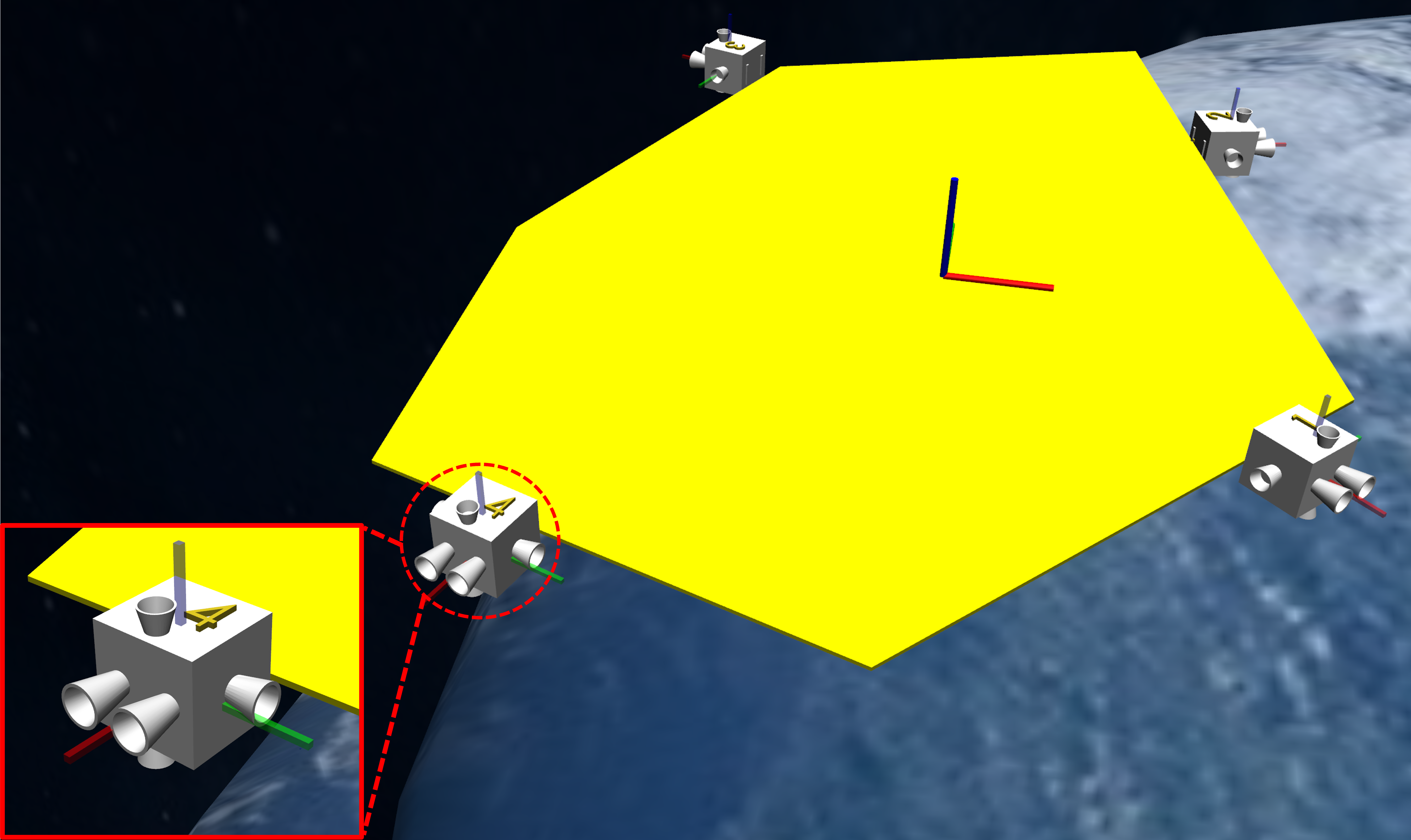}
  \vspace{0.6cm}
  \caption{ }
  \label{mujoco_b}
\end{subfigure}%
\begin{subfigure}{.5\textwidth}
  \centering
  \includegraphics[width=0.97\linewidth]{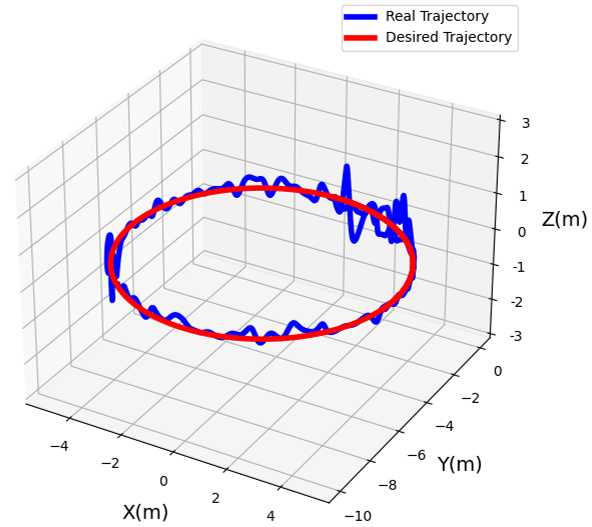}
   \caption{ }
  \label{sim2-a}
\end{subfigure}\\
\vspace{0.2cm}
  \begin{subfigure}{.5\textwidth}
  \centering
  \includegraphics[width=0.95\linewidth]{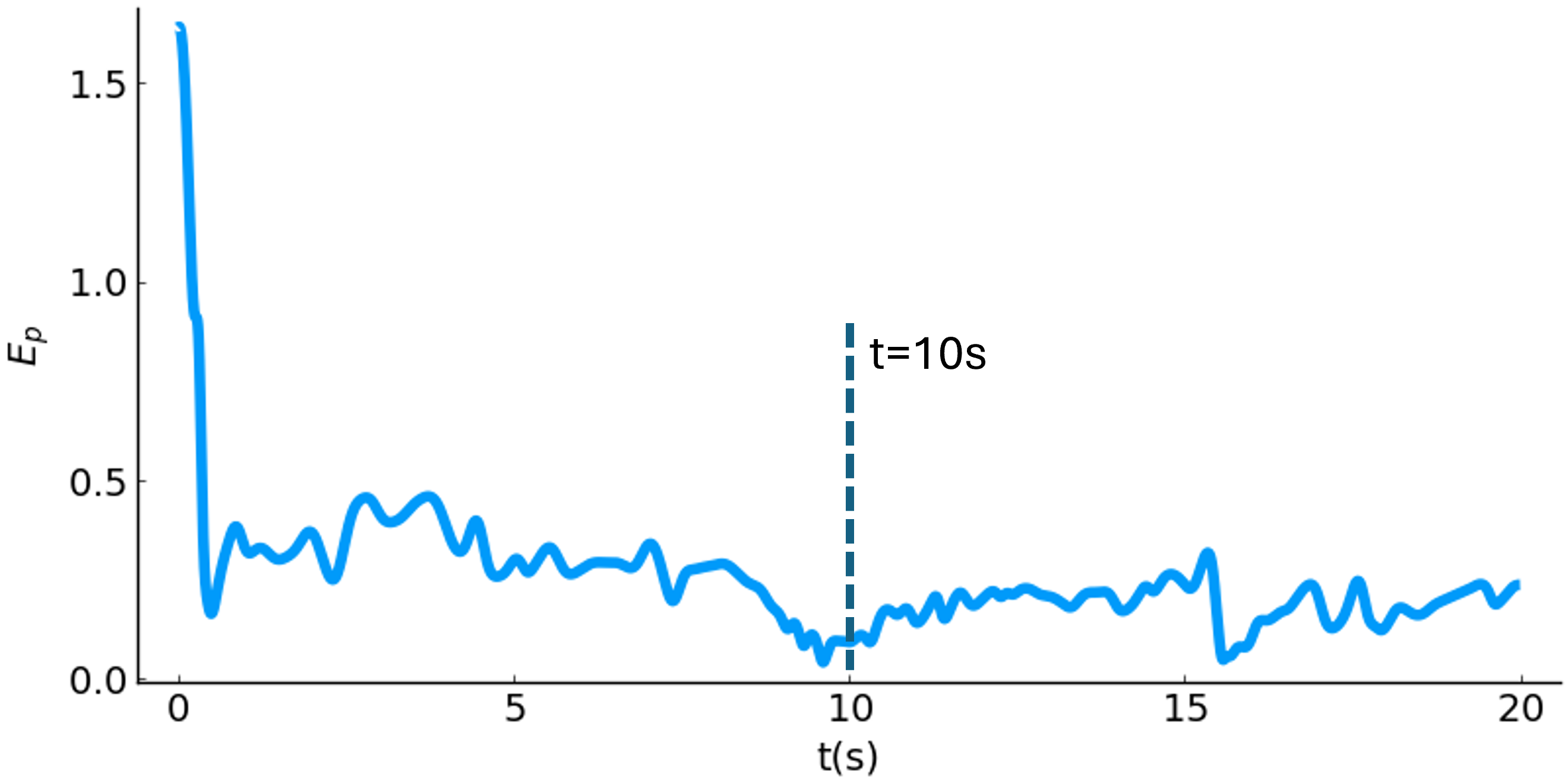}
  \caption{ }
  \label{sim2-b}
\end{subfigure}%
\begin{subfigure}{.5\textwidth}
  \centering
  \includegraphics[width=0.95\linewidth]{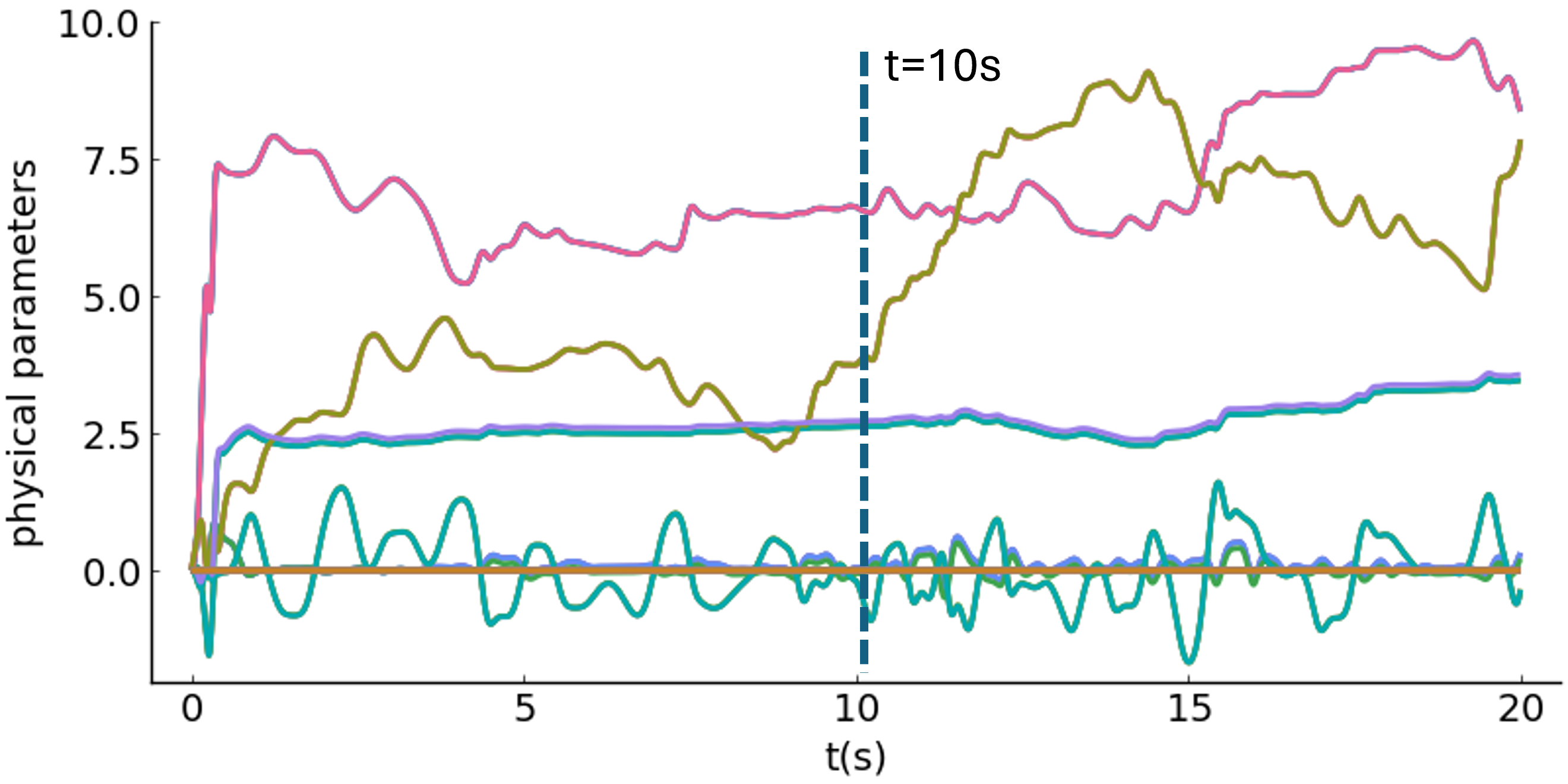}
   \caption{ }
  \label{sim2-c}
\end{subfigure}\\
\vspace{0.2cm}
\begin{subfigure}{.5\textwidth}
  \centering
  \includegraphics[width=0.95\linewidth]{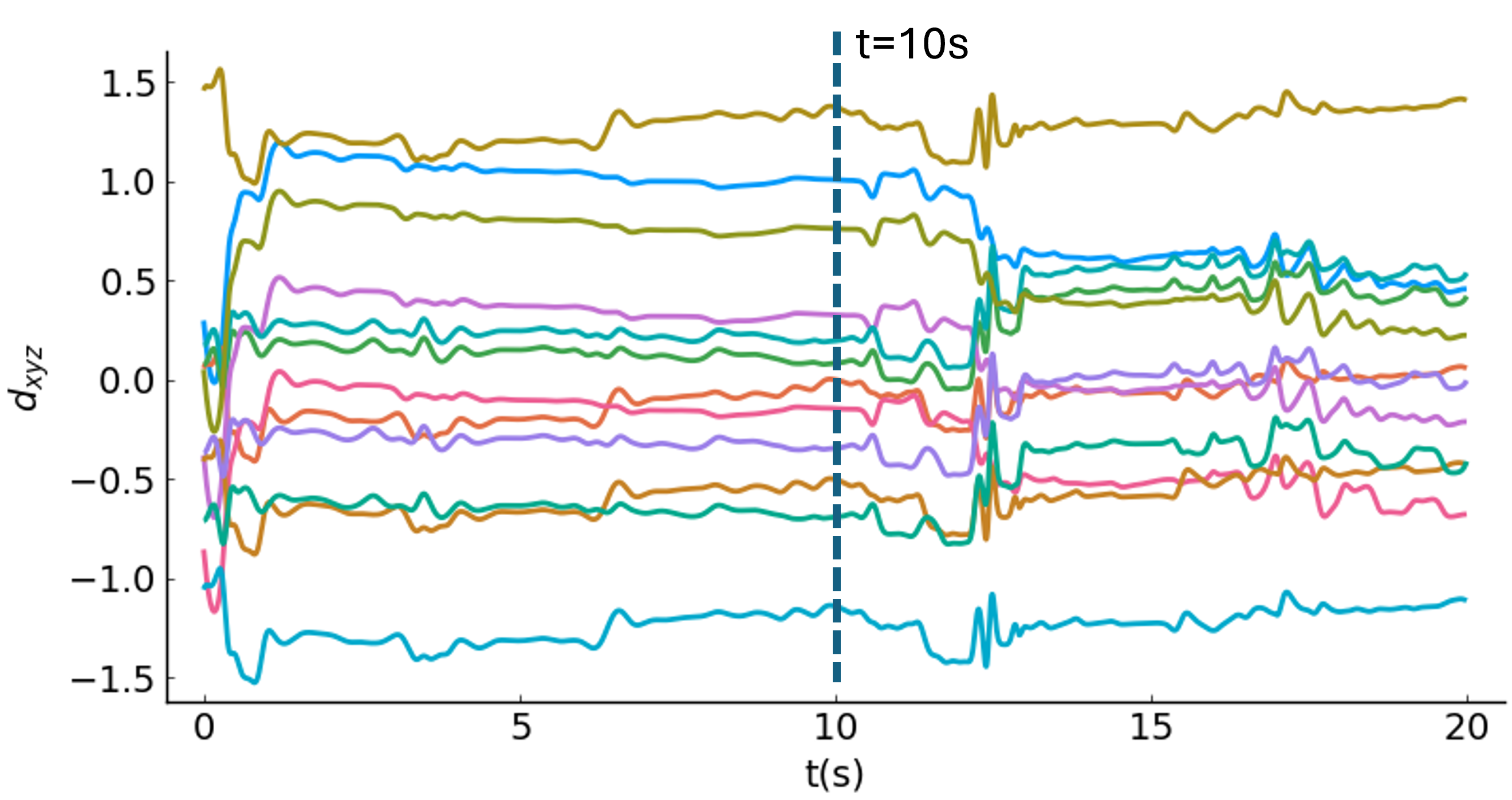}
  \caption{ }
  \label{sim2-d}
\end{subfigure}%
\begin{subfigure}{.5\textwidth}
  \centering
  \includegraphics[width=0.95\linewidth]{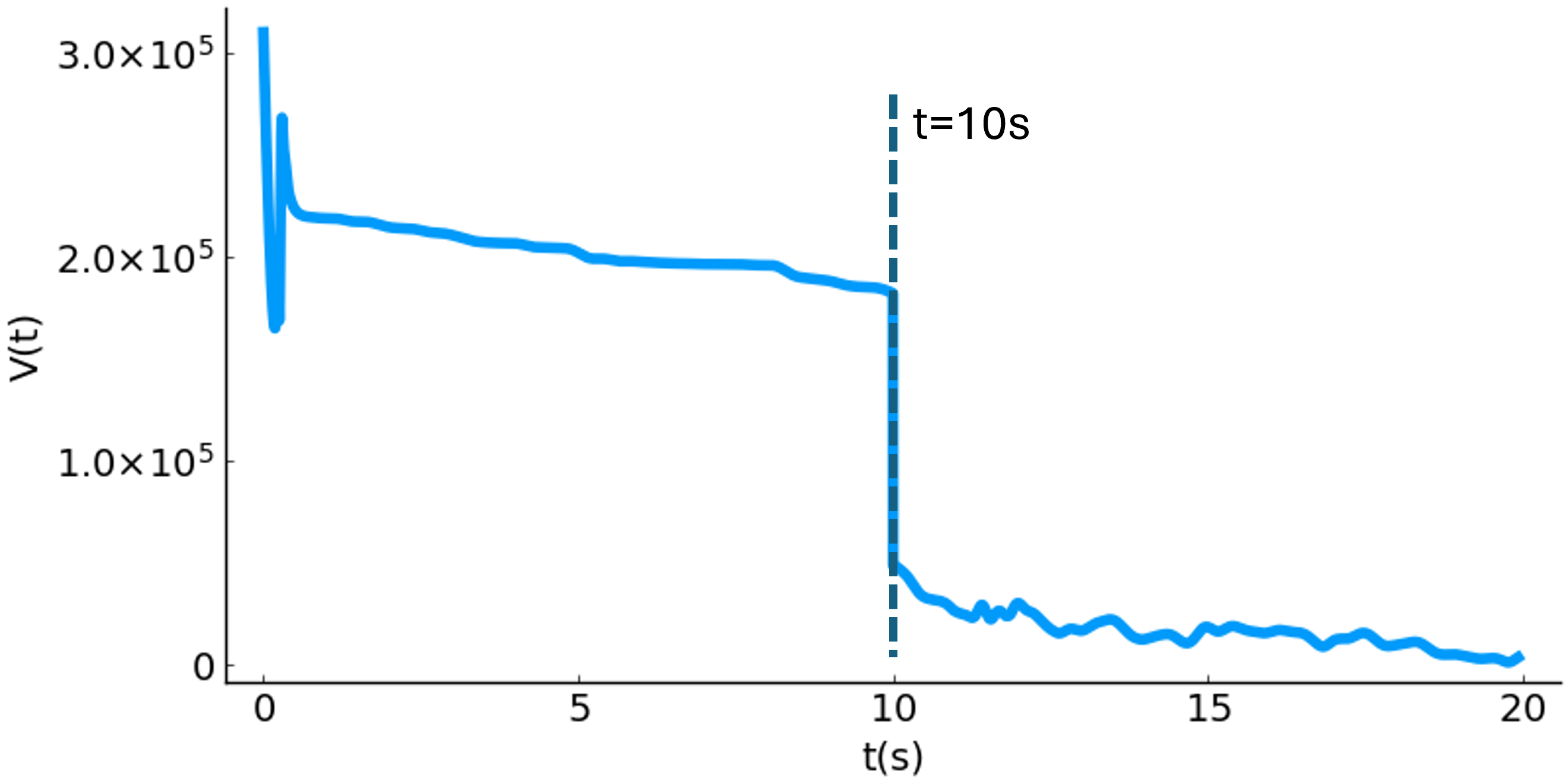}
   \caption{ }
  \label{sim2-e}
\end{subfigure}
\caption{(a) Transportation task implemented on a group of space tugs in space on MuJoCo platform. (b)-(f) Simulation results for using $n$ space tugs collaboratively to transport the payload in $SE(3)$ under zero-gravity environment.}
\end{figure}

Fig. \ref{sim2-a} shows the tracking effect over time in 3D space. Fig. \ref{sim2-b} shows the composite error~\eqref{eq:error} for the client manipulated by a group of space tugs in the space environment. The results show at $10$ s, we disable tug-1, which introduces the disturbance to the system.

Fig. \ref{sim2-c} and Fig. \ref{sim2-d} show the parameter estimation for the physical properties of the client and the contact point for each tug to the measurement point $\boldsymbol{P}_s$. Based on the results, we can see that all signals are bounded, and the final values are close to the actual values.

Fig. \ref{sim2-e} shows the Lyapunov-like function $V(t)$ versus time for the simulation results. The Lyapunov-like function is positive $V(t) > 0$ and its derivative $\dot{V}(t) \leq 0$. This verifies that our system is stable in the sense of Lyapunov.

%\begin{figure}[H]
 % \centering 
  %\captionsetup[sub]{font=scriptsize,labelfont={sf},oneside,margin={0.75cm,0cm}}
 % \subfloat[]{\includegraphics[width=2.in]{Results/3D_TPOD_v1.png} \label{sim2-a}}\\ 
 % \qquad 
  %\subfloat[]{\includegraphics[width=2.in]{Results/TPOD_pos_error_v1.png}\label{sim2-b}} 
  %\qquad 
  %\subfloat[]{\includegraphics[width=2.in]{Results/TPOD_physic_est_v1.png}\label{sim2-c}} 
  %\qquad 
  %\subfloat[]{\includegraphics[width=2.in]{Results/TPOD_distance_est_v1.png}\label{sim2-d}} 
  %\qquad 
  %\subfloat[]{\includegraphics[width=2.2in]{Results/Lyapunov_dis_TPOD_v1.png}\label{sim2-e}} 
  %\caption{Simulation results for using $n$ space tugs collaboratively to transport the payload in $SE(3)$ under zero-gravity environment.}
  %\label{fig:figure1_2}
%\end{figure}

% \begin{figure}
%      \centering
%     \begin{subfigure}[t]{0.34\textwidth}
%         \raisebox{-\height}{\includegraphics[width=\textwidth]{Results/3D_TPOD_v1.png}}
%     \caption{} 
%     \end{subfigure}
%     \hfill
%     \begin{subfigure}[t]{0.65\textwidth}
%         \raisebox{-\height}{\includegraphics[width=0.48\textwidth]{Results/TPOD_pos_error_v1.png}}
%         \label{b}
%         \raisebox{-\height}{\includegraphics[width=0.48\textwidth]{Results/TPOD_physic_est_v1.png}}%
%         \vspace{2ex}
%         \raisebox{-\height}{\includegraphics[width=0.48\textwidth]{Results/TPOD_distance_est_v1.png}}
%         \raisebox{-\height}{\includegraphics[width=0.48\textwidth]{Results/Lyapunov_dis_TPOD_v1.png}}
%         \caption{}
%     \end{subfigure}
%     \caption{caption of main figure}
% \end{figure}

\section{Conclusion and Future Work}
We addressed the problem of decentralized control in transportation tasks by using multiple aerospace vehicles. All vehicles cooperatively manipulate and transport a rigid object without prior knowledge of its mass, inertia, center of mass, and grasping points. We presented a novel decentralized adaptive controller design, implemented it on multiple fully actuated hexarotor UAVs in a gravitational  environment, and then extended it to multiple space tugs for the same transportation task to follow the desired position and orientation in a zero-gravity environment. We proved the feasibility of our controller in two simulation environments. During the transportation mission, we deactivated one of the vehicles to introduce a disturbance to the whole system and examine the robustness of our method. Two simulations using $n = 4$ fully actuated hexarotor UAVs in an Earth environment and $n = 4$ space tugs in a space environment, along with related results, demonstrated the efficacy of the presented controller design. 

We believe that the improvements made in the design of the decentralized adaptive control can be applied to other critical robotic servicing tasks where the assignment and distribution of agents is crucial. These tasks include having multiple robots working collaboratively in manufacturing processes. We are currently working on learning-based robust adaptive control~\cite{zhang2022robust}, implemented on multiple drones for transporting and dislodging non-rigid objects with unknown physical properties, including mass, inertia, and material flexibility, to handle servicing tasks using multiple drones in a wide range of applications.

\section{Acknowledgement}
This material is based on research sponsored by Air Force Research Laboratory (AFRL) under agreements FA9453-18-2-0022 and FA9550-22-1-0093. We would like to acknowledge Zhi Zheng from Zhejiang University for useful discussion in dynamics analysis of the drone and also acknowledge Giovanni Cordova for his assistance with Solidworks modeling.

\newpage
\bibliography{main}

\begin{thebibliography}{10}

\bibitem{gao2024adaptive}
Longsen Gao, Claus Danielson, and Rafael Fierro.
\newblock Adaptive robot detumbling of a non-rigid satellite.
\newblock {\em arXiv preprint arXiv:2407.17617}, 2024.

\bibitem{down2023adaptive}
Ian Down and Manoranjan Majji.
\newblock Adaptive detumbling of uncontrolled planar spacecraft using finite module deposition.
\newblock In {\em AIAA SCITECH 2023 Forum}, page 0158, 2023.

\bibitem{gao2023autonomous}
Longsen Gao, Giovanni Cordova, Claus Danielson, and Rafael Fierro.
\newblock Autonomous multi-robot servicing for spacecraft operation extension.
\newblock In {\em 2023 IEEE/RSJ International Conference on Intelligent Robots and Systems (IROS)}, pages 10729--10735. IEEE, 2023.

\bibitem{hert2023mrs}
Daniel Hert, Tomas Baca, Pavel Petracek, Vit Kratky, Robert Penicka, Vojtech Spurny, Matej Petrlik, Matous Vrba, David Zaitlik, Pavel Stoudek, et~al.
\newblock Mrs drone: A modular platform for real-world deployment of aerial multi-robot systems.
\newblock {\em Journal of Intelligent \& Robotic Systems}, 108(4):64, 2023.

\bibitem{saldana2018modquad}
David Saldana, Bruno Gabrich, Guanrui Li, Mark Yim, and Vijay Kumar.
\newblock Modquad: The flying modular structure that self-assembles in midair.
\newblock In {\em 2018 IEEE International Conference on Robotics and Automation (ICRA)}, pages 691--698. IEEE, 2018.

\bibitem{salinas2023unified}
Lucio~R Salinas, Javier Gimenez, Daniel~C Gandolfo, Claudio~D Rosales, and Ricardo Carelli.
\newblock Unified motion control for multilift unmanned rotorcraft systems in forward flight.
\newblock {\em IEEE Transactions on Control Systems Technology}, 2023.

\bibitem{cardona2021adaptive}
Gustavo~A Cardona, Diego~S D'Antonio, Rafael Fierro, and David Salda{\~n}a.
\newblock Adaptive control for cooperative aerial transportation using catenary robots.
\newblock In {\em 2021 Aerial Robotic Systems Physically Interacting with the Environment (AIRPHARO)}, pages 1--8. IEEE, 2021.

\bibitem{sun2023nonlinear}
Sihao Sun and Antonio Franchi.
\newblock Nonlinear mpc for full-pose manipulation of a cable-suspended load using multiple uavs.
\newblock In {\em 2023 International Conference on Unmanned Aircraft Systems (ICUAS)}, pages 969--975. IEEE, 2023.

\bibitem{wang2024enhancing}
Xinrui Wang and Yan Jin.
\newblock Enhancing efficiency in collision avoidance: A study on transfer reinforcement learning in autonomous ships’ navigation.
\newblock {\em ASME Open Journal of Engineering}, 3, 2024.

\bibitem{10611458}
Weihan Wang, Chieh Chou, Ganesh Sevagamoorthy, Kevin Chen, Zheng Chen, Ziyue Feng, Youjie Xia, Feiyang Cai, Yi~Xu, and Philippos Mordohai.
\newblock Stereo-nec: Enhancing stereo visual-inertial slam initialization with normal epipolar constraints.
\newblock In {\em 2024 IEEE International Conference on Robotics and Automation (ICRA)}, pages 2691--2697, 2024.

\bibitem{wang2023model}
Han Wang, Leonardo~F Toso, Aritra Mitra, and James Anderson.
\newblock Model-free learning with heterogeneous dynamical systems: A federated lqr approach.
\newblock {\em arXiv preprint arXiv:2308.11743}, 2023.

\bibitem{bosio2023automated}
Carlo Bosio, Jerry Tang, Ting-Hao Wang, and Mark~W Mueller.
\newblock Automated layout design and control of robust cooperative grasped-load aerial transportation systems.
\newblock {\em arXiv preprint arXiv:2310.07649}, 2023.

\bibitem{mu2019universal}
Bingguo Mu and Pakpong Chirarattananon.
\newblock Universal flying objects: Modular multirotor system for flight of rigid objects.
\newblock {\em IEEE Transactions on Robotics}, 36(2):458--471, 2019.

\bibitem{barawkar2024decentralized}
Shraddha Barawkar, Manish Kumar, and Michael Bolender.
\newblock Decentralized adaptive controller for multi-drone cooperative transport with offset and moving center of gravity.
\newblock {\em Aerospace Science and Technology}, 145:108960, 2024.

\bibitem{chaikalis2023modular}
Dimitris Chaikalis, Nikolaos Evangeliou, Anthony Tzes, and Farshad Khorrami.
\newblock Modular multi-copter structure control for cooperative aerial cargo transportation.
\newblock {\em Journal of Intelligent \& Robotic Systems}, 108(2):31, 2023.

\bibitem{mellinger2013cooperative}
Daniel Mellinger, Michael Shomin, Nathan Michael, and Vijay Kumar.
\newblock Cooperative grasping and transport using multiple quadrotors.
\newblock In {\em Distributed Autonomous Robotic Systems: The 10th International Symposium}, pages 545--558. Springer, 2013.

\bibitem{icra18gripper}
Bruno Gabrich, David Saldaña, Vijay Kumar, and Mark Yim.
\newblock A flying gripper based on cuboid modular robots.
\newblock In {\em 2018 IEEE International Conference on Robotics and Automation (ICRA)}, pages 7024--7030, 2018.

\bibitem{li2023wrench}
Zhongmou Li, Vincent B{\'e}goc, Abdelhamid Chriette, and Isabelle Fantoni.
\newblock Wrench capability analysis and control allocation of a collaborative multi-drone grasping robot.
\newblock {\em Journal of Mechanisms and Robotics}, 15(2):021003, 2023.

\bibitem{flores2022fully}
Alejandro Flores and Gerardo Flores.
\newblock Fully actuated hexa-rotor uav: Design, construction, and control. simulation and experimental validation.
\newblock In {\em 2022 International Conference on Unmanned Aircraft Systems (ICUAS)}, pages 1497--1503. IEEE, 2022.

\bibitem{hexa2015franchi}
S.~{Rajappa}, M.~{Ryll}, H.~H. {Bülthoff}, and A.~{Franchi}.
\newblock Modeling, control and design optimization for a fully-actuated hexarotor aerial vehicle with tilted propellers.
\newblock In {\em 2015 IEEE International Conference on Robotics and Automation (ICRA)}, pages 4006--4013, 2015.

\bibitem{ollero2022past}
Anibal Ollero, Marco Tognon, Alejandro Suarez, Dongjun Lee, and Antonio Franchi.
\newblock Past, present, and future of aerial robotic manipulators.
\newblock {\em IEEE Transactions on Robotics}, 38(1):626--645, 2022.

\bibitem{khamseh2018aerial}
Hossein~Bonyan Khamseh, Farrokh Janabi-Sharifi, and Abdelkader Abdessameud.
\newblock Aerial manipulation—a literature survey.
\newblock {\em Robotics and Autonomous Systems}, 107:221--235, 2018.

\bibitem{yang2022collaborative}
Chenyu Yang, Guo~Ning Sue, Zhongyu Li, Lizhi Yang, Haotian Shen, Yufeng Chi, Akshara Rai, Jun Zeng, and Koushil Sreenath.
\newblock Collaborative navigation and manipulation of a cable-towed load by multiple quadrupedal robots.
\newblock {\em IEEE Robotics and Automation Letters}, 7(4):10041--10048, 2022.

\bibitem{culbertson2021decentralized}
Preston Culbertson, Jean-Jacques Slotine, and Mac Schwager.
\newblock Decentralized adaptive control for collaborative manipulation of rigid bodies.
\newblock {\em IEEE Transactions on Robotics}, 37(6):1906--1920, 2021.

\bibitem{aghili2019robust}
Farhad Aghili.
\newblock Robust impedance-matching of manipulators interacting with uncertain environments: Application to task verification of the space station's dexterous manipulator.
\newblock {\em IEEE/ASME Transactions on Mechatronics}, 24(4):1565--1576, 2019.

\bibitem{khalil2002nonlinear}
K~Hassan Khalil.
\newblock {\em Nonlinear systems}.
\newblock Prentice Hall, 2002.

\bibitem{slotine1987adaptive}
Jean-Jacques~E Slotine and Weiping Li.
\newblock On the adaptive control of robot manipulators.
\newblock {\em The international journal of robotics research}, 6(3):49--59, 1987.

\bibitem{todorov2012mujoco}
Emanuel Todorov, Tom Erez, and Yuval Tassa.
\newblock Mujoco: A physics engine for model-based control.
\newblock In {\em 2012 IEEE/RSJ International Conference on Intelligent Robots and Systems}, pages 5026--5033. IEEE, 2012.

\bibitem{stoer1996introduction}
Josef Stoer and Roland Bulirsch.
\newblock {\em Introduction to Numerical Analysis}, volume~12.
\newblock Texts in Applied Mathematics, Springer, New York, 1996.

\bibitem{zhang2022robust}
Xinglong Zhang, Jiahang Liu, Xin Xu, Shuyou Yu, and Hong Chen.
\newblock Robust learning-based predictive control for discrete-time nonlinear systems with unknown dynamics and state constraints.
\newblock {\em IEEE Transactions on Systems, Man, and Cybernetics: Systems}, 52(12):7314--7327, 2022.

\end{thebibliography}
\end{document}